\documentclass[dvipsnames,table]{article}

\usepackage[final]{hunyuan}

\usepackage{amsmath}
\usepackage{amssymb}
\usepackage{mathtools}
\usepackage{amsthm}
\usepackage{multirow}
\usepackage{makecell}
\usepackage{subcaption}
\usepackage{wrapfig}
\usepackage{graphicx}

\usepackage{pifont}
\usepackage{fvextra}
\usepackage{xspace}

\usepackage[utf8]{inputenc} 
\usepackage[T1]{fontenc}    
\usepackage{url}            
\usepackage{booktabs}       
\usepackage{amsfonts}       
\usepackage{nicefrac}       
\usepackage{microtype}      
\usepackage{colortbl}
\usepackage{color, soul}
\usepackage[table]{xcolor}
\usepackage{booktabs}
\usepackage{textgreek}

\usepackage[labelfont=bf]{caption}
\usepackage{enumitem}
\usepackage{sectsty}
\usepackage{pgfplots}
\pgfplotsset{compat=1.17}
\usepackage{fancyhdr}
\usepackage{natbib} 
\let\cite\citep    

\renewcommand{\headrulewidth}{1pt}
\makeatletter
\def\headrule{{\if@fancyplain\let\headrulewidth\plainheadrulewidth\fi
\hrule\@height\headrulewidth\@width\textwidth \vskip-\headrulewidth}}
\makeatother

\definecolor{HYDarkBlue}{HTML}{2155EA} 
\definecolor{HYLightBlue}{HTML}{A8DFF6}
\definecolor{HYLightBlueLighter}{HTML}{C6ECFA}

\definecolor{syh}{RGB}{41, 91, 160}
\usepackage[colorlinks, linkcolor=HYDarkBlue, anchorcolor=HYDarkBlue, citecolor=HYDarkBlue, urlcolor=HYDarkBlue]{hyperref}

\usepackage{graphicx}   
\usepackage{booktabs}   
\usepackage{amsmath}    
\usepackage{natbib}     
\usepackage{cleveref}   

\usepackage{threeparttable}

\sectionfont{\color{HYDarkBlue}\fontfamily{zi4}\selectfont}
\subsectionfont{\color{HYDarkBlue}\fontfamily{zi4}\selectfont}
\subsubsectionfont{\color{HYDarkBlue}\fontfamily{zi4}\selectfont}

\title{DFly2: Turbocharging Long-Context Speculative Decoding by Sharing the Target's Sparsity}

\author{%
\textbf{\normalsize{Hong Liu,\quad Rui Cen,\quad Junhan Shi,\quad Lin Niu,\quad Runzhi Fan,\quad Zhaojing Zhou,\quad Haodi Lei,}} \\
\textbf{\normalsize{Jiebin Zhang,\quad Saiyong Yang,\quad Junqiang Zheng,\quad Zhenghai Xue,\quad Guanghua Yu \textsuperscript{*},\quad Jianchen Zhu}}\\[0.4em]
{\normalfont\small Tencent Inc.}\\
{\normalfont\small \textsuperscript{*}Corresponding author}
\vspace{0em}
}

\def\github{\raisebox{-1.5pt}{\includegraphics[height=1.05em]{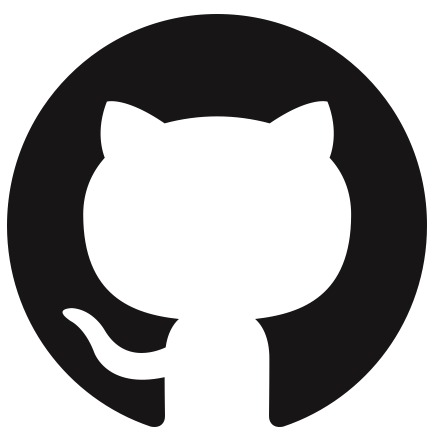}}}
\def\email{\raisebox{-1.5pt}{\includegraphics[height=1.05em]{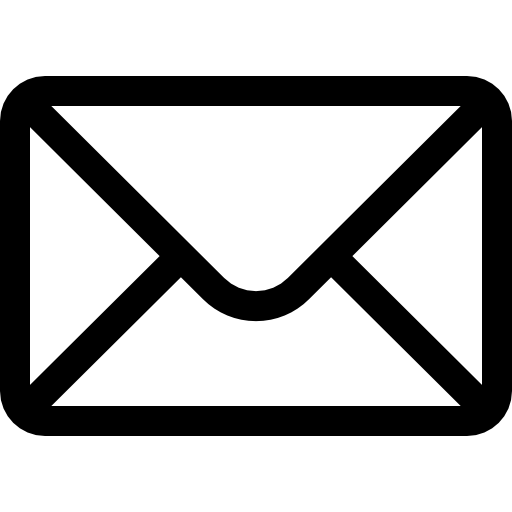}}}

\usepackage{geometry}
\usepackage{algorithm}
\usepackage{algpseudocode}
\usepackage{setspace}

\makeatletter
\renewcommand{\ALG@beginalgorithmic}{\small}
\makeatother



\begin{document}

\maketitle
\thispagestyle{fancy} 

\begin{abstract}
To avoid the quadratic cost of dense attention over long contexts, models such as DeepSeek-V4 and Hy4
have turned natively sparse.
Speculative decoding adds a further, lossless speedup with fewer target passes, and matters most
for the long contexts these models serve, where decoding is slowest.
Block-parallel drafters, however, still work in a paradigm built for dense targets, keeping a context
cache of their own over the whole prompt or a sliding window.
The full one takes a larger share of the KV per token the more the target compresses its own, and
either costs more than memory, in prefill time, prefix-cache hits and decode concurrency.
Drafters that read the target's KV cache avoid one of their own, but work on this line settles
only which cache, not which positions, and still reads every position or a recent window.
Neither rule is made for the query at hand, and computing a selection per query in the drafter would
bring back a cost that grows with the context.
Yet a natively sparse target already picks positions for each query, over a cache it keeps anyway.
Both of the drafter's choices are therefore already made, and paid for, by the target: its cache and
its selection.
We measure how far one selection carries across a block: made a step early, it still covers close to
nine tenths of what each later token selects on DS-V4.1-Flash.
We therefore present \textbf{DFly2, a block-parallel drafter that shares the target's sparsity}.
Each draft layer reads the KV its paired target layer cached (\textbf{KV sharing}), at the positions
that layer's indexer selected, a fixed number from anywhere in the context (\textbf{Index sharing}).
DFly2 runs no scoring pass and keeps no cache of its own, so none of those costs apply.
We evaluate on three targets whose sparse attention and architecture differ, DS-V4.1-Flash,
GLM-5.3-Flash and Hy4~flash, over general-purpose, long-context and agentic benchmarks at two
temperatures.
Across all of them, DFly2 accepts more tokens per step than DFlash and DSpark trained on the same
recipe.
Its margin over the stronger baseline widens from \textbf{3.9\%--10.5\%} on general workloads to
\textbf{10.0\%--31.5\%} on long-context and agentic workloads of up to 128K.
Serving DS-V4.1-Flash at high concurrency and matched per-user decode speed, it delivers \textbf{62\%}
and \textbf{81\%} more throughput than that baseline on long-context and agentic sets at 64K.
\end{abstract}

\begin{figure*}[h]
\centering
\includegraphics[width=\linewidth]{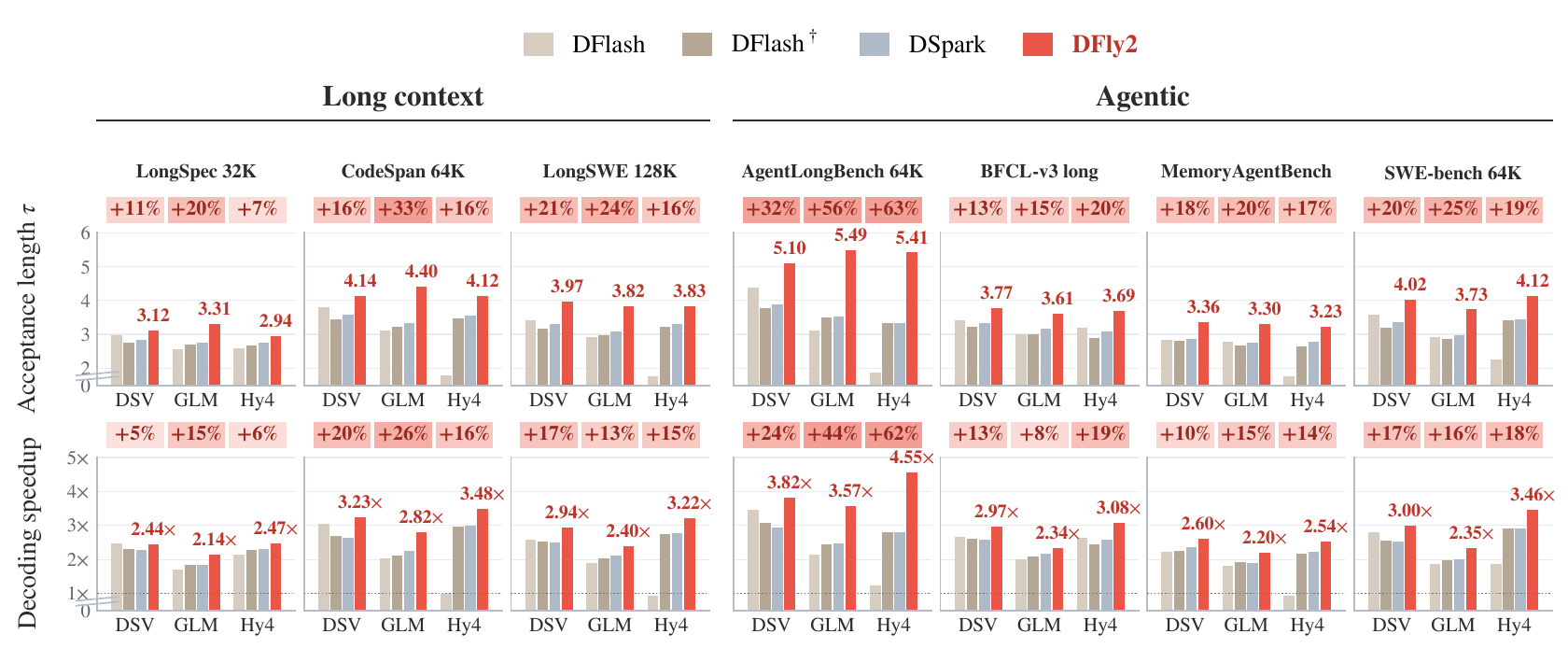}
\caption{Accepted length $\tau$ and decoding speedup on DS-V4.1-Flash, GLM-5.3-Flash and
Hy4~flash, greedy at a concurrency of one. Chips give DFly2's gain over the better of DFlash$^\dagger$
and DSpark.}
\label{fig:teaser}
\end{figure*}

\section{Introduction}
\label{sec:intro}

Agents and multi-turn chat have pushed serving to long contexts, where the cost of dense attention grows
quadratically as the context lengthens.
Recent long-context models such as DeepSeek-V4, GLM-5 and Hy4 have answered by turning natively
sparse~\citep{nsa,deepseekv32,deepseekv4,glm5,hy4}.
Each sparse layer attends only to a fixed budget of positions, which a trained indexer picks for the current
query.
They also store less, compressing the KV cache they keep for the
context~\citep{deepseekv32,deepseekv4,glm5,hy4}.
At a million tokens, DeepSeek-V4-Flash needs 10\% of the per-token FLOPs and 7\% of the KV cache of
DeepSeek-V3.2~\citep{deepseekv4}.

Speculative decoding, a mainstream lossless accelerator, speeds up decoding further with fewer target passes,
and serving needs it most in agentic and multi-turn workloads, whose long contexts slow decoding
down~\citep{leviathan,chen2023,magicdec,dspark}.
Long-context speculative decoding thus increasingly runs on natively sparse targets: DeepSeek-V3.2,
DeepSeek-V4, GLM-5 and Hy4 all ship with a drafter~\citep{deepseekv4,dspark,glm5,hy4}.
Sparsity has made each decode step of such a target cheaper, but drafters still work in a paradigm designed
for dense targets.
We focus on block-parallel drafters, which make two choices about the target's context: where it comes from,
meaning which cache and which positions, and how it is injected~\citep{dflash}.
The mainstream answer takes the context from a cache of the drafter's own, over positions picked by a rule
that ignores the current query~\citep{dflash,dflare,hyperdflash,angelspec,dspark,treeflash,domino}.
That cache holds keys and values the drafter projects from the target's hidden states and injects into every
draft layer.
DFlash caches every position, and this cache weighs more the more the target saves (\Cref{sec:obs:own}).
It adds a fifth to the KV per token on Hy4~flash and three quarters on DeepSeek-V4.1-Flash
(DS-V4.1-Flash).
DSpark bounds that cost with a sliding window, which on DeepSeek-V4 spans the last 128 tokens of the
context~\citep{dspark}.

Full or windowed, a cache of the drafter's own costs more than the memory it takes.
The drafter has to run on the prefill instances under prefill-decode disaggregation and build the cache over
the prompt, which delays the first token (\Cref{sec:obs:own,sec:exp:results}).
A full cache also leaves the target fewer and shorter prefix-cache hits, and removing it lets DS-V4.1-Flash
hold 1.73$\times$ as many requests at 32K tokens (\Cref{sec:obs:own}).

Drafters that read the target's KV cache instead avoid a growing cache of their own, a line that goes back to
GliDe~\citep{glide} and runs through LongSpec, H-Spec and LongSpark~\citep{longspec,hspec,longspark}.
They settle which cache to read, but not which positions.
Like DFlash and DSpark, these drafters still read every position or a recent window, by rules that ignore the
current query because the dense targets they were developed on offered no selection to follow
(\Cref{sec:related}).
Attention over all positions also grows with the context, while a fixed number of positions costs the same
at every length (\Cref{sec:obs:own}).
For drafters with a cache of their own, the better rule even depends on the target.
On long-context and agentic workloads, a full cache beats a window on DS-V4.1-Flash, loses to one on
GLM-5.3-Flash, and collapses on Hy4~flash once prompts pass the 32K the drafter was trained at
(\Cref{sec:exp:results}).
A rule carried over from dense targets thus needs to be adapted to each target.
Each of these rules is fixed when the drafter is designed, not made for the query at hand, and scoring the
context instead would bring back a cost that grows with it.
The cache a drafter reads can be the target's own, so could the selection it follows come from the target too?

A natively sparse target already makes that selection at every step: its trained indexers pick a fixed budget
of positions for the current query, from a KV cache the target already keeps.
This is the selection that drafting has wanted and given up: top-$k$ attention ``can achieve very high
acceptance rate'', yet ``it is not a practical draft option because of its prohibitively high KV selection
cost''~\citep{magicdec}.
Verification-guided methods recover it by extracting it from a dense target's
attention~\citep{sparsespec,vegas,specextend}, a step a natively sparse target does not need.
Our key insight is simple: a natively sparse target has already settled where a drafter's context should come
from, with its cache and its selection, and paid for both.
One selection also holds across a block: made one step early, for the token just before it, it serves every
slot.
On DS-V4.1-Flash it covers close to nine tenths of what each later token would select, against less than a
fifth for a window of the same size (\Cref{sec:obs:early}).

We present \textbf{DFly2, a block-parallel drafter that shares the target's sparsity}: the compact KV cache
it maintains and the positions its sparse layers attend to.
Each draft layer reads the KV that a paired target layer has cached (\textbf{KV sharing}), at the positions
that layer's indexer selected, a fixed number drawn from anywhere in the context (\textbf{Index sharing}).
DFly2 thus runs no scoring pass and keeps no cache of its own, so it needs no place on the prefill instances,
inherits every prefix-cache hit of the target, and costs no decode concurrency (\Cref{sec:method}).
The selection was made for the target's queries, yet the drafter is a far smaller model with queries of its
own.
Each draft layer starts from its paired layer's weights, including the query latent that layer shares with its
indexer, so draft queries begin in the space that produced the selection.
With the target's last-layer hidden state, which no cache holds, gated into the block, DFly2 reuses four
things the target has already computed: its KV, its selection, its layer weights and that state.
An intra-block correction then rereads the shared context slot by slot.

We instantiate DFly2 on three targets that differ in their variant of DeepSeek Sparse Attention
(DSA)~\citep{deepseekv32} and in architecture, taking each target's selection in the form it gives
(\Cref{sec:method}).
DS-V4.1-Flash keeps one cache shared across its layers as YOCO does~\citep{yoco}, GLM-5.3-Flash interleaves
DSA layers with linear attention ones, and Hy4~flash compresses the index of its sparse layers in two stages
(\Cref{sec:bg}).
On all three, DFly2 accepts more tokens per step than DFlash and DSpark trained on the same recipe, on every
benchmark and temperature (\Cref{sec:exp:results}).
Its margin over the stronger baseline grows from 3.9\%--10.5\% on general workloads to 10.0\%--31.5\% on
long-context and agentic ones.
On agentic workloads, whose prompts mostly run longer, its average stays within 4\% of its general average,
while DSpark's falls by 11\% to 21\%.
The target's selection reaches anywhere in the context yet stays a fixed size, so DFly2 inherits its length
generalization.
What DFly2 reads is not tuned to a target but taken from it, which is why the same design holds on all three
architectures.
Accepted length turns into speed, and with nothing added at prefill, DFly2 averages a higher end-to-end
speedup than the baselines on every target and workload family (\Cref{sec:exp:serving}).

Our contributions are threefold:

\begin{itemize}
\item \textbf{A drafter can take the target's selection as it is.}
      A context of the drafter's own costs more than memory, reaching prefill, prefix caching and decode
      concurrency.
      Our insight is that the selection it needs is already computed, and paid for, by the target.
      We measure how far one selection carries: made one step early, it still covers close to nine tenths of
      what each later token in the block would select on DS-V4.1-Flash.

\item \textbf{DFly2 shares the target's selection alongside its cache.}
      DFly2 takes both: each draft layer reads its paired target layer's KV cache, at the positions that
      layer's indexer selected, and starts from that layer's weights.
      DFly2 therefore keeps no cache, runs no scoring pass, and reads a fixed number of positions from
      anywhere in the context, so the costs above do not apply.
      We add two pieces: the last-layer state that no cache holds, and an intra-block correction that rereads
      the shared context.

\item \textbf{Gains that grow with the context, on all three targets.}
      Against baselines trained on the same recipe, DFly2's margin in accepted length over the stronger one
      grows from 3.9\%--10.5\% on general workloads to 10.0\%--31.5\% on long-context and agentic ones.
      It generalizes past its training length: on agentic workloads its average accepted length stays within
      4\% of its general average.
      Serving at high concurrency on DS-V4.1-Flash, at matched per-user decode speed, it delivers 28\% more
      throughput on short text and 62\% and 81\% more on a long-context and an agentic set at 64K.
\end{itemize}


\providecommand{\resultrows}{\renewcommand{\arraystretch}{1.2}}

\section{Background}
\label{sec:bg}

\paragraph{Speculative decoding.}
At each step a drafter proposes $\gamma$ tokens and the target verifies them in one forward pass, accepting
the longest prefix consistent with its distribution and appending one token of its own, the bonus
token~\citep{leviathan,chen2023}.
With $\tau$ the mean accepted length, which counts the bonus token and so satisfies $1 \le \tau \le \gamma+1$,
the latency per generated token is
\begin{equation}
\label{eq:sd-latency}
T_{\mathrm{tok}} \;=\; \frac{T_{\mathrm{draft}} + T_{\mathrm{verify}}}{\tau}.
\end{equation}
Drafting time in turn splits into a context-independent part $T_0$, spent in projections, feed-forward
layers and the output head, and attention over the $m$ context positions whose KV the drafter reads,
\begin{equation}
\label{eq:draft-split}
T_{\mathrm{draft}} \;=\; T_0 + T_{\mathrm{attn}}(m).
\end{equation}
Speedup thus comes from raising $\tau$ while keeping $T_{\mathrm{draft}}$ small, and $T_{\mathrm{draft}}$
grows with the context only through $m$.

\paragraph{Block-parallel drafters.}
A block-parallel drafter fills a whole block in one forward pass~\citep{dflash}.
Its input is an anchor token followed by $\gamma$ mask tokens, the draft slots, which attend to one another
and to the context.\footnote{We call positions rows, as in the target's attention matrix. The anchor, the
previous round's bonus token, sits at row $A$, and the target generated it from its query at row $A-1$.
Draft slot $k$ predicts the token at row $A+k$ and so stands in for the target's query at row $A+k-1$.}
Whatever its architecture, a drafter is conditioned on the target's context, which reaches it in one of
three ways (\Cref{tab:routes}).
Autoregressive drafters such as MTP~\citep{mtp} and EAGLE~\citep{eagle3} take target hidden states at
their input and keep a KV cache of their own over the whole context.
The other two routes put the context into the keys and values of every draft layer $i$,
\begin{equation}
\label{eq:kv-injection}
K_i = \bigl[\, X_i \;;\; W_i^{K} H_{\mathrm{blk}} \,\bigr],
\qquad
V_i = \bigl[\, Y_i \;;\; W_i^{V} H_{\mathrm{blk}} \,\bigr],
\end{equation}
where $H_{\mathrm{blk}}$ holds the block's hidden states and $[\,\cdot\,;\,\cdot\,]$ stacks along the
sequence.
KV injection sets $(X_i, Y_i) = (W_i^{K} H_{\mathrm{ctx}}, W_i^{V} H_{\mathrm{ctx}})$, with $H_{\mathrm{ctx}}$
fused and projected from the hidden states of several target layers~\citep{dflash,dspark}.
Reading the target's cache instead builds them from the latents the target has already stored for a paired
layer~\citep{glide}.
The first two routes thus leave the drafter with KV of its own for every context position, and the third
adds none.

\begin{table*}[t]
\centering
\small
\caption{\textbf{How a drafter receives the target's context.} Only reading the target's cache leaves the
drafter without a context KV of its own.}
\label{tab:routes}
\vspace{4pt}
\resultrows
\begin{tabular}{@{}lll@{}}
\toprule
\textbf{Route} & \textbf{Context enters as} & \textbf{Drafter's own context KV} \\
\midrule
Input conditioning (MTP, EAGLE)    & target hidden states at the input       & every position \\
KV injection (DFlash, DSpark)      & $H_{\mathrm{ctx}}$ in every layer's K/V & every position \\
Reading the target's cache (GliDe) & the target's stored K/V, read in place  & none \\
\bottomrule
\end{tabular}
\end{table*}

\paragraph{Natively sparse targets.}
The three targets we study are natively sparse: each is trained with a variant of DeepSeek Sparse Attention
(DSA)~\citep{deepseekv32}, which recent long-context models have widely adopted~\citep{deepseekv4,glm5}.
In each sparse attention layer, a lightweight indexer gives every preceding token $s$ an index score
$I_{t,s}$ for the query token $\mathbf{h}_t$, and the layer attends only to the key-value latents
$\mathbf{c}_s$ that score highest to produce the output $\mathbf{u}_t$,
\begin{equation}
\label{eq:dsa}
\mathbf{u}_t = \operatorname{Attn}\bigl(\mathbf{h}_t,\ \{\mathbf{c}_s \mid s \in \mathcal{S}_t\}\bigr),
\qquad
\mathcal{S}_t = \bigl\{\, s \mid I_{t,s} \in \operatorname{Top-k}(I_{t,:}) \,\bigr\}.
\end{equation}
The selection $\mathcal{S}_t$ is made afresh for every query and has a fixed size, while the scoring pass
still runs over the whole context~\citep{spin}.
These models also shrink what they store: most layers keep only a sliding window or a recurrent state, and
the three differ in the remaining layers.
DS-V4.1-Flash~\citep{deepseekv4} stores a compressed latent in four of its 40 layers and shares it across
the others as YOCO does~\citep{yoco},
GLM-5.3-Flash~\citep{glm5} interleaves 11 DSA layers with 34 linear attention layers, and
Hy4~flash~\citep{hy4flash} compresses the index of its sparse layers in two stages.
The KV that grows with the context is therefore small, and both this compact layout and the selection
$\mathcal{S}_t$ are part of the target itself.


\providecommand{\resultrows}{\renewcommand{\arraystretch}{1.2}}

\section{Observations}
\label{sec:obs}

A drafter makes two choices about its context, whether to keep its own KV cache and which positions to
read, and both point to the target.
Its own cache costs it at prefill, in prefix caching and at decode, and sharing the target's KV avoids all
three (\Cref{sec:obs:own}).
At long contexts it should read only a bounded number of positions, and the target already selects a fixed
number of them for its own attention (\Cref{sec:obs:which}).
Made one step before the block, this selection still covers most of what later tokens would select
(\Cref{sec:obs:early}).

\subsection{Should the drafter keep its own KV cache?}
\label{sec:obs:own}

Owning a cache of its own costs a drafter in three places, and attention over the context adds a fourth cost at every step.

\paragraph{Prefill.}
At prefill, a cache of its own adds compute but mainly constrains where the drafter runs.
Under prefill-decode disaggregation the drafter must also run on the prefill instances, where it holds its
weights, projects an injection at every prompt position and transfers the resulting cache to the decode
instance.
KV sharing adds nothing at prefill, because the latents the drafter reads are ones the target has already written.

\paragraph{Prefix caching.}
A cache of its own also leaves the target fewer and shorter prefix-cache hits.
The drafter's own cache is projected from the target's hidden states, which prefix caching does not keep.
It must therefore either be kept in the prefix cache or be rebuilt by recomputing the target.
If kept, it takes over two fifths of each cached prefix's memory on DS-V4.1-Flash
(\Cref{tab:pertoken-kv}), so the pool holds fewer prefixes for multi-turn and agentic workloads to reuse.
Rebuilding it shortens the target's own hit, since the target must recompute the hidden states the hit
would have skipped.
After each hit, vLLM~\citep{vllm} recomputes the last 128 tokens to restore the target's sliding windows on
DS-V4.1-Flash, so a drafter window avoids both only if it is no longer than those 128 tokens.
Under KV sharing the drafter has no cache of its own to keep or rebuild, so every target hit is also a drafter hit.

\begin{table*}[h]
\centering
\small
\caption{\textbf{A drafter's own cache costs most where the target has saved most.} Per-token KV of all
caches that grow with the context, from model configurations (bf16), for a three-layer drafter.}
\label{tab:pertoken-kv}
\vspace{4pt}
\resultrows
\begin{tabular}{@{}lrcr@{}}
\toprule
\textbf{Target} & \textbf{Target KV} & \textbf{Drafter's own cache} & \textbf{Increase} \\
\midrule
DS-V4.1-Flash & 4.4 KiB  & 3.4 KiB & \textbf{1.76$\times$} \\
GLM-5.3-Flash       & 11.7 KiB & 3.0 KiB & 1.26$\times$ \\
Hy4~flash           & 17.2 KiB & 3.4 KiB & 1.20$\times$ \\
\bottomrule
\end{tabular}
\end{table*}
\begin{figure}[t]
\centering
\begin{minipage}{0.49\linewidth}
  \centering
  \includegraphics[width=\linewidth]{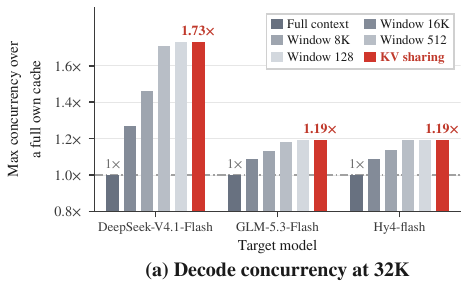}
\end{minipage}\hfill
\begin{minipage}{0.49\linewidth}
  \centering
  \includegraphics[width=\linewidth]{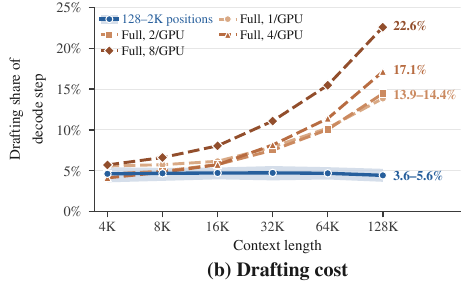}
\end{minipage}
\caption{\textbf{What a drafter's own cache costs at decode.}
(a) Concurrency at 32K tokens, with a full own cache as 1$\times$, estimated from model configurations
including fixed per-request state.
(b) DS-V4.1-Flash, timed per step by the server, with the band measured as a window.}
\label{fig:system-cost}
\end{figure}

\paragraph{Decode.}
A drafter that keeps its own cache for the whole context, as DFlash does, lowers decode concurrency.
Every in-flight request's KV stays resident during decode, so a pool of $M$ bytes holds at most
$M/(bL+f)$ requests of length $L$.
Here $b$ is the per-token KV that grows with the context, including the drafter's own cache if it keeps
one, and $f$ the fixed per-request state.
The concurrency lost is largest on the targets that have saved the most, because their layer design
shrinks the target's KV but not the drafter's own cache.
Even in the target's compact latent layout, three draft layers add three quarters to $b$ on
DS-V4.1-Flash (\Cref{tab:pertoken-kv}).
Removing that addition, as KV sharing does, lets the same pool hold 1.73$\times$ as many requests at 32K
tokens (\Cref{fig:system-cost}\,a).
A sliding-window drafter~\citep{dspark} trades reach for concurrency: the closer it comes to that ratio,
the less of the context its window covers (\Cref{fig:system-cost}\,a).
KV sharing attains that ratio without a window, so the drafter can read anywhere in the context.

\paragraph{Drafting.}
The last cost comes from attention over the context rather than from owning a cache.
By \Cref{eq:draft-split}, drafting time grows with the context only through $T_{\mathrm{attn}}(m)$, whose
compute and memory traffic scale with the number $m$ of positions read.
A fixed budget of positions, as a window or a top-$k$ selection, costs 4\% to 6\% of the decode step at
every context length and load.
The full cache costs as much at 4K tokens but climbs to 14\% to 23\% at 128K, faster under heavier load
(\Cref{fig:system-cost}\,b).
This cost should follow the number of positions, wherever they lie and whichever cache holds them.
A drafter that shares the target's KV but reads all of it then pays that rising cost too.
At long contexts a drafter should therefore read a bounded number of positions, and the most useful ones
cost as little as the most recent.

\begin{table*}[t]
\centering
\small
\caption{\textbf{KV sharing removes the drafter's own cache without limiting its reach.}
DS-V4.1-Flash, $w = 128$ or 512, concurrency estimated as in \Cref{fig:system-cost}\,a.
$^{\dagger}$Except at $w = 128$.}
\label{tab:escapes}
\vspace{4pt}
\resultrows
\begin{tabular}{@{}lccc@{}}
\toprule
 & \multicolumn{2}{c}{\textbf{Drafter's own cache}} & \\
\cmidrule(lr){2-3}
 & \textbf{Full} & \textbf{Window $w$} & \textbf{KV sharing} \\
\midrule
Drafter needed at prefill & yes, for all positions  & yes, for the last $w$        & \textbf{no} \\
Target's prefix-cache hits & fewer and shorter      & can be cut short$^{\dagger}$ & \textbf{unaffected} \\
Decode concurrency at 32K & 1.00$\times$            & 1.71--1.73$\times$           & \textbf{1.73$\times$} \\
Reach at 32K              & \textbf{whole context}  & last $w$ (0.4--1.6\%)        & \textbf{whole context} \\
\bottomrule
\end{tabular}
\end{table*}

\medskip
\noindent
A sliding-window drafter shrinks its own cache but still has to run on the prefill side
(\Cref{tab:escapes}).
It recovers most of the concurrency and keeps prefix caching intact only by giving up reach.
Trained with a 512- or 128-token window, DFlash loses 3\% to 12\% of its full-context accepted length on
long contexts, the most at the 128 tokens that prefix caching requires.
Sharing the target's KV removes the drafter's own cache and, by design, all three costs of owning it, while
keeping the whole context within reach.
With the read bounded as well, the drafter's cost should no longer grow with the context, and the main
question left is which positions to read without losing that reach.

\subsection{Which positions should the drafter read?}
\label{sec:obs:which}

A bounded read can pick its positions by position, which yields a window of the most recent
ones~\citep{dspark} and costs reach (\Cref{sec:obs:own}).
It can instead pick by query, reaching the whole context at the price of a scoring pass over all positions.
That pass grows with the context like a full read, giving back what the bound saved~\citep{magicdec}.

A natively sparse target already pays for this scoring pass.
At every step its indexers select a fixed budget of positions for its sparse attention layers, 512 to
2{,}048 in released models.\footnote{512 and 1{,}024 on DeepSeek-V4-Flash and
V4-Pro~\citep{deepseekv4}, 2{,}048 on DeepSeek-V3.2~\citep{deepseekv32} and on GLM-5.3-Flash.}
A drafter that shares the target's KV can share the selection as well, reading by query at the cost of a
window and with no scoring of its own (\Cref{fig:system-cost}\,b).
The drafter is trained to predict what the target would generate, so we expect it to need the same context
the target attends to.

NSA~\citep{nsa} lends indirect support: a selection trained with the model outperforms one computed
afterwards, and the target's indexers were trained with the target.
Following the target's selection still rests on two assumptions.
The drafter is a different model, so learning to use the selection is a matter of training that
\Cref{sec:exp} tests end to end.
The selection, made one step before the block, must still cover what the rows in the block would select,
which \Cref{sec:obs:early} measures on the target alone, with no drafter involved.

\begin{figure}[t]
\centering
\includegraphics[width=\linewidth]{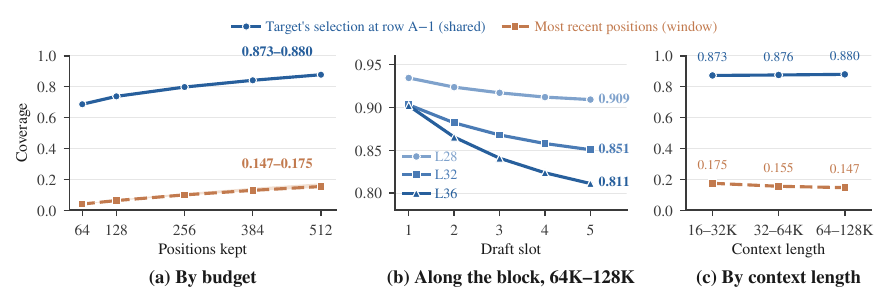}
\caption{Top-$k$ mass coverage on DS-V4.1-Flash. (a) Mean over slots and layers. Bands span context
lengths. (b) Budget 512, own scale. The window stays at 0.13--0.17. (c) Budget 512, averaged as in (a).}
\label{fig:coverage}
\end{figure}

\subsection{Does the target's selection hold across the block?}
\label{sec:obs:early}

\paragraph{Question.}
The second assumption is that a selection made for one query serves the queries that follow.
The target selects one query at a time, so a drafter starting a block has only the selection made for the
token just before the anchor.
The drafter applies that selection to every token in the block.
Each later token, however, has a query of its own, which depends on tokens the target has not yet processed
and may favor other positions.

\paragraph{Measurement.}
In the notation of \Cref{sec:bg}, the shared selection is $\mathcal{S}_{A-1}$, and draft slot $k$ stands in
for row $A+k-1$.
We run the target on the actual continuation to obtain each row's own top-512, $\mathcal{S}_t$ of
\Cref{eq:dsa}.
With $p_t$ the index-score distribution of row $t$, the top-$k$ mass coverage
\begin{equation}
\label{eq:coverage}
\mathrm{Cov}(t) \;=\; \sum_{s \in \mathcal{S}_{A-1}} p_t(s) \;\Big/\; \sum_{s \in \mathcal{S}_t} p_t(s)
\end{equation}
is the share of what row $t$ would select that the shared selection already holds.
Its denominator is the most any 512 positions can hold, so $\mathrm{Cov}(t) \le 1$.
We measure it on DS-V4.1-Flash over 158{,}638 blocks of long-context evaluation data.

\paragraph{Findings.}
The selection largely holds.
Averaged over slots and layers, the coverage of \Cref{eq:coverage} reaches close to nine tenths, while a
window of the same size recovers less than a fifth at the same cost (\Cref{fig:coverage}\,a).
Along the block it declines only gradually, and the deepest layer's last slot still keeps about four
fifths (\Cref{fig:coverage}\,b), well above what the window reaches at any slot.
Its margin over the window widens as the context grows (\Cref{fig:coverage}\,c), because the target spreads
its picks over the whole context while the window covers less and less of it.
A selection the target has already paid for therefore serves the whole block, not only the row it was made
for.

\paragraph{Mechanism.}
The gradual decline points to the query's content rather than its position.
Rotating the query at row $A-1$ to each later row leaves its top-512 unchanged in all 72 slot, layer and
context cells we checked.
The energy of an index query sits almost entirely in its low-frequency RoPE components, which a shift of a
few positions barely moves.
That leaves the content, which changes one token at a time, so the target's focus drifts rather than
switches.
The decline should therefore follow how far into the block a row sits rather than how long the context
is.

\subsection{Summary}
\label{sec:obs:summary}

Both choices behind a drafter's context lead back to the target.
Owning a cache of its own costs a drafter at prefill, in prefix caching and in decode concurrency, three
costs that sharing the target's KV avoids by design without giving up reach (\Cref{sec:obs:own}).
Long contexts also call for a bounded read, and a natively sparse target has already paid for a query-aware
selection once judged too costly for drafting~\citep{magicdec} (\Cref{sec:obs:which}).
Made one step early, that selection largely holds across the block and covers far more than a window of the
same size (\Cref{sec:obs:early}).
\Cref{sec:method} turns both into mechanisms, \emph{KV sharing} to read the target's KV cache and
\emph{Index sharing} to read the positions the target itself selected.
The drafter then keeps no cache and runs no indexer of its own, and can still read anywhere in the
context.

\section{Method}
\label{sec:method}

\begin{figure*}[t]
\centering
\includegraphics[width=0.75\linewidth]{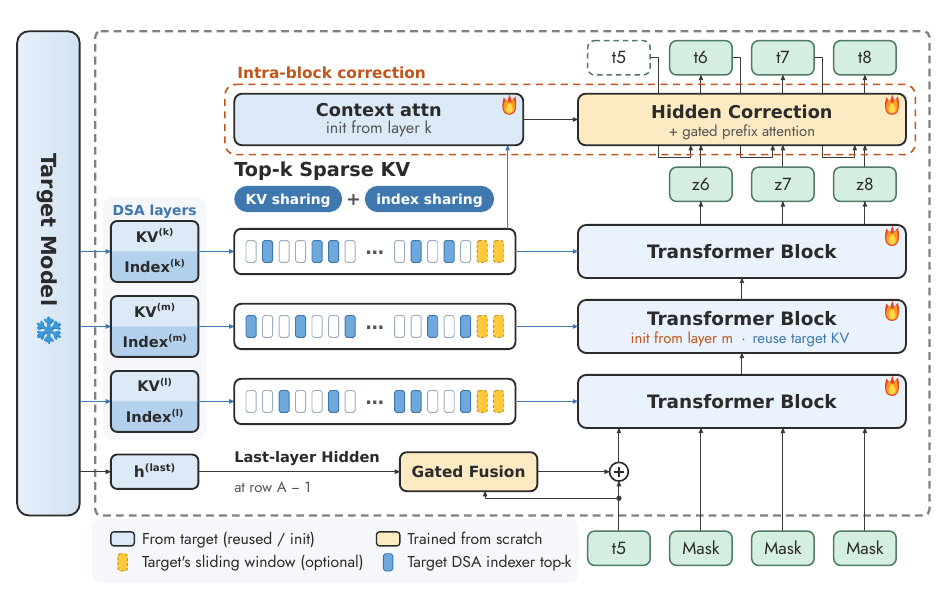}
\caption{Overview of DFly2. Each of its three draft layers is paired with one of the target's DSA layers
and reads the KV that layer cached, at the positions its indexer selected.}
\label{fig:method}
\end{figure*}

A drafter must decide where its context comes from and which positions to read (\Cref{sec:obs}).
DFly2 leaves both decisions to the target (\Cref{fig:method}).
Its three layers fill one block in a single pass, each paired with one target layer.
A draft layer reads the KV its paired layer has already cached, at the positions that layer selected.

\begin{figure}[t]
\centering
\includegraphics[width=\linewidth]{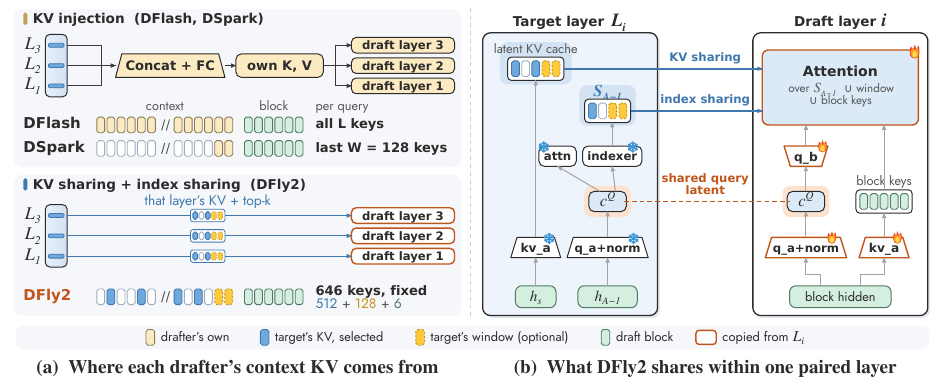}
\caption{(a) DFlash and DSpark build their own KV, while DFly2 reads the target's KV at its top-$k$
positions. (b) Only the latent KV and the selection pass at run time, so the drafter needs no indexer.}
\label{fig:sharing}
\end{figure}

\paragraph{KV sharing.}
The context a draft layer attends to is the target's (\Cref{fig:sharing}a).
Before the block, the KV injection of DFlash and DSpark projects every position into keys and values the
drafter stores (\Cref{eq:kv-injection})~\citep{dflash,dspark}.
DFly2 instead reads the latents the paired layer has already cached,
\begin{equation}
\label{eq:kv-sharing}
(X_i,\, Y_i) \;=\; \bigl(\, \underbrace{W^{UK}_{i}}_{\text{trained}}\,
\underbrace{\mathbf{C}^{L_i}_{\mathcal{R}_i}}_{\text{shared}} \,,\;\;
W^{UV}_{i}\, \mathbf{C}^{L_i}_{\mathcal{R}_i} \,\bigr),
\qquad
\mathbf{C}^{L_i}_{\mathcal{R}_i} \;=\; \{\mathbf{c}^{L_i}_{s}\}_{s \in \mathcal{R}_i},
\end{equation}
where $\mathbf{c}^{L_i}_{s}$ is the latent that layer $L_i$ holds at position $s$ and $\mathcal{R}_i$ is the
set of positions draft layer $i$ reads, itself taken from the target (\Cref{eq:index-sharing}).
Inside the block the drafter still compresses its $\gamma$ slots with a down-projection $W^{D}_{i}$
and expands them with the same up-projections, so a draft layer's keys and values read
\begin{equation}
\label{eq:kv-full}
K_i \;=\; W^{UK}_{i}\bigl[\, \underbrace{\mathbf{C}^{L_i}_{\mathcal{R}_i}}_{\text{context}} \;;\;
\underbrace{W^{D}_{i} H_{\mathrm{blk}}}_{\text{the block}} \,\bigr],
\qquad
V_i \;=\; W^{UV}_{i}\bigl[\, \mathbf{C}^{L_i}_{\mathcal{R}_i} \;;\; W^{D}_{i} H_{\mathrm{blk}} \,\bigr],
\end{equation}
with $H_{\mathrm{blk}}$ the hidden states of the block and $W^{UK}_{i} W^{D}_{i}$ the factored form of
$W^{K}_{i}$ in \Cref{eq:kv-injection}.
Both terms are latents, so one attention covers the context and the block at once.
What the drafter shares at run time is the target's cache, not its weights.
The target supplies $\mathbf{C}^{L_i}_{\mathcal{R}_i}$, while every projection in
\Cref{eq:kv-sharing,eq:kv-full} belongs to the drafter and is trained (\Cref{fig:sharing}b).
Only the target's DSA layers keep reusable latents, so the $L_i$ are drawn from them in depth order, one
per draft layer.
These layers do not select the same positions, so the drafter sees several views of the context rather
than one view repeated (\Cref{tab:targets} lists $L$ and \Cref{sec:exp:ablation} ablates the choice).

The drafter stores nothing of its own per context position.
In \Cref{eq:kv-sharing,eq:kv-full} the context comes from the target's cache, and the rest is projection
matrices of fixed size.
The three costs of \Cref{sec:obs:own} therefore do not apply: the drafter needs no place on the prefill
instances, leaves the target's prefix-cache hits unaffected, and costs no decode concurrency.
A short window shrinks these costs without removing them.

\paragraph{Index sharing.}
A draft layer reads where its paired layer reads, and $\mathcal{R}_i$ in \Cref{eq:kv-sharing} is that
set:
\begin{equation}
\label{eq:index-sharing}
\mathcal{R}_i \;=\; \mathcal{S}^{(L_i)}_{A-1} \;\cup\; \mathcal{N},
\qquad
m \;=\; |\mathcal{R}_i|.
\end{equation}
Here $\mathcal{S}^{(L_i)}_{A-1}$ is the top-$k$ the indexer of layer $L_i$ scored at row $A-1$
(\Cref{eq:dsa}), and $\mathcal{N}$ the target's near window before the anchor.
What the second term adds follows from the target's structure.
One such set serves all $\gamma$ slots of a layer, and a selection made one step early largely holds
across the block (\Cref{sec:obs:early}).

The drafter neither scores the context nor reads more of it as the context grows.
The verify pass that produced the anchor already made the selection, so a draft layer runs no scoring
pass of its own and carries no indexer (\Cref{fig:sharing}b).
Both terms are fixed in size by the target, so the number of positions a draft step reads, $m$, does not
grow with the context (\Cref{fig:sharing}a).
A draft step's attention over the context is therefore bounded, and so is the drafting time $T_{\mathrm{draft}}$
(\Cref{eq:draft-split}).

The two terms differ across the three targets, and a draft layer takes them as each target gives them,
with nothing to tune (\Cref{tab:targets}).
The first term is a top-$k$ over single positions on DS-V4.1-Flash and over groups of adjacent
positions on GLM-5.3-Flash, so the same $k$ can mean several times as many positions.
Where the target stores its near window apart from the latents the top-$k$ picks from, as
DS-V4.1-Flash does, the second term adds what the top-$k$ cannot pick.
On GLM-5.3-Flash and Hy4~flash the near window lies within those latents, so the top-$k$ can already
pick it (\Cref{sec:exp:ablation} ablates the term).
\Cref{sec:exp:analysis} asks whether reading beyond $\mathcal{R}_i$ would help.

\paragraph{Initializing from the target.}
The paired layer also gives a draft layer its starting weights, so one pairing settles what the draft
layer reads, where it reads and what it starts from.
A draft layer's \textbf{attention} has the form of the paired layer's and starts from that layer's
weights,
\begin{equation}
\label{eq:attn-init}
O_i \;=\; W^{O}_{i} \operatorname{Attn}\bigl(
\underbrace{W^{UQ}_{i}\, \mathrm{Norm}(W^{DQ}_{i} H_{\mathrm{blk}})}_{Q_i},\;
K_i,\; V_i \bigr),
\qquad
W^{\star}_{i}\big|_{\mathrm{init}} \;=\; W^{\star}_{L_i},
\end{equation}
with $K_i$ and $V_i$ from \Cref{eq:kv-full} and $\star$ running over $DQ$, $UQ$, $D$, $UK$, $UV$ and $O$.
The norms are copied with the matrices, and all of them are trained from there.
Orthrus likewise starts its trainable projections from the target's~\citep{orthrus}.
On a natively sparse target $W^{DQ}_{L_i}$ also feeds the indexer queries, so the copy inherits the
target's \emph{shared query latent} even without the indexer (\Cref{fig:sharing}b).
A draft query therefore starts in the query space that produced the top-$k$ it reuses, on all three
targets.

The \textbf{feed-forward} side also starts from the paired layer, but a drafter is small by design, so a
draft layer draws only a fraction of that layer's routed experts,
\begin{equation}
\label{eq:ffn-init}
\theta^{\mathrm{exp}}_{i}\big|_{\mathrm{init}} \;=\;
\theta^{\mathrm{exp}}_{L_i}\bigl[\, \mathcal{E}_i \,\bigr],
\qquad
\mathcal{E}_i \;\sim\; \mathrm{Unif}\bigl\{\, \mathcal{B} \subseteq \mathcal{E}^{L_i} \;:\;
|\mathcal{B}| = E \,\bigr\},
\end{equation}
where $\theta^{\mathrm{exp}}$ is a layer's routed experts, $\mathcal{E}^{L_i}$ their ids at layer $L_i$,
and $E$ how many a draft layer holds.
Only the choice of experts is random, and their weights are the target's.
Each drawn expert keeps its router row and correction bias, and the shared expert and the
post-attention norm are copied whole.
\Cref{sec:exp:ablation} measures each half of the initialization alone and the two together.

\paragraph{Last-layer hidden injection.}
The target's last-layer hidden state goes only to its LM head and MTP layer, so no cache holds it and
reading the target's KV does not reach it.
Yet it is the state the target produced the anchor from, and MTP heads draft from it~\citep{mtp}.
DFly2 fuses it with the anchor's embedding, as the target's MTP layer does, and adds the result at the
anchor slot through a gate that starts at zero (\Cref{fig:method}).
In a disaggregated deployment this adds one vector per request to the prefill-to-decode transfer.

\begin{figure}[t]
\centering
\includegraphics[width=\linewidth]{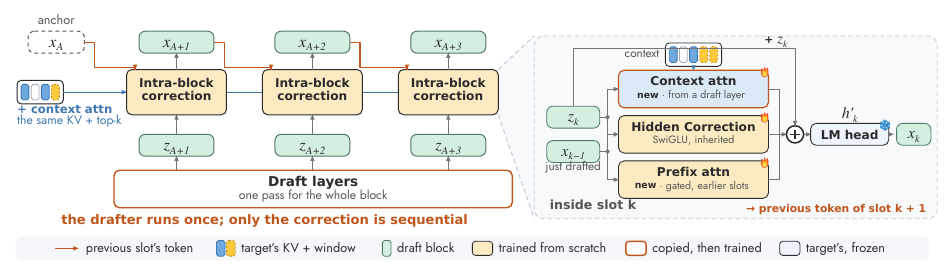}
\caption{Intra-block correction across the block (left) and inside slot $k$ (right). Slot $k$ is corrected
from its hidden state and the token drafted at slot $k-1$.}
\label{fig:correction}
\end{figure}

\paragraph{Intra-block correction.}
The correction DFly2 runs inside the block returns to the context.
Block-parallel drafters correct slot $k$ from its hidden state and the token drafted at slot $k-1$.
TreeFlash's SwiGLU over that pair is the \emph{hidden correction} we inherit unchanged~\citep{treeflash}.
Such an update can only rework what the slot already holds: after \emph{the value in row 3 is} it cannot
recover the number, and that is where corrections elsewhere stop (\Cref{sec:related}).
Reading the context would normally mean storing or projecting something per position, and here it costs
neither, since the context is the target's and already in its cache.
Alongside the inherited update, a \emph{context attention} reads one draft layer's context and a
\emph{gated prefix attention} the block up to slot $k$ (\Cref{fig:correction}),
\begin{equation}
\label{eq:correction}
\mathbf{h}_k \;\leftarrow\; \mathbf{h}_k \;+\;
\underbrace{\mathrm{SwiGLU}(\mathbf{z}_k)}_{\text{inherited}} \;+\;
W^{O}_{c}\bigl[\, \underbrace{\operatorname{Attn}\bigl(\mathbf{q}_k,\,
\mathbf{C}^{L_i}_{\mathcal{R}_i}\bigr)}_{\text{context}} \;+\;
\underbrace{\mathbf{g}_k \odot \operatorname{Attn}\bigl(\mathbf{q}_k,\,
\mathbf{h}_{0:k}\bigr)}_{\text{the block}} \,\bigr],
\end{equation}
where $i$ is the draft layer it reads, $\mathbf{z}_k = [\, \mathrm{Norm}(\mathbf{h}_k) \,;\,
\mathrm{Norm}(\mathrm{Emb}(y_{k-1})) \,]$ is that same pair, and the query $\mathbf{q}_k$ and the per-head
gate $\mathbf{g}_k$ are projected from it.
The correction's own output projection $W^{O}_{c}$ starts at zero with $\mathbf{g}_k$, so it begins as the
inherited one alone.
It runs slot by slot, the one sequential part of a draft step (\Cref{sec:exp:ablation} weighs its cost
against what it buys).

\paragraph{Training.}
Only the drafter's parameters are trained, and \Cref{sec:exp:setup} gives the data and the schedule.
The objective is single-stage: each slot is matched to the target's distribution there, and the slots
are weighted by the accepted length that depends on them,
\begin{equation}
\label{eq:loss}
\mathcal{L} \;=\; \frac{\sum_{k=1}^{\gamma} w_k\, \ell_k}{\sum_{k=1}^{\gamma} w_k},
\qquad
\ell_k \;=\; \lambda_k \operatorname{KL}\bigl(p_k \,\|\, \hat{p}_k\bigr) \;+\;
(1-\lambda_k) \operatorname{TV}\bigl(p_k,\, \hat{p}_k\bigr),
\end{equation}
where $p_k$ is the target's distribution at slot $k$, $\hat{p}_k$ the drafter's, and $\ell_k$ the LK
distillation loss~\citep{lk}, whose $\lambda_k$ follows the overlap between the two.
The D-PACE weights $w_k$~\citep{dpace} are the accepted length still expected from slot $k$ on, taken
from the drafter's probabilities at the earlier slots and carrying no gradient.
Serving takes the selection from the verify pass, and training, which has none, recomputes it with a
frozen copy of each paired layer's indexer.
A draft layer therefore sees the same latents and the same positions in training as at serving.

\definecolor{dflyrow}{RGB}{238,246,252}
\newcommand{\grouprule}{\arrayrulecolor{black!40}\midrule[\cmidrulewidth]\arrayrulecolor{black}}
\newlength{\grouppad}
\setlength{\grouppad}{\dimexpr(\aboverulesep+\belowrulesep+\cmidrulewidth)/8\relax}
\newcommand{\grouprulepad}{\noalign{\vskip\grouppad}\grouprule\noalign{\vskip\grouppad}}
\providecommand{\tightrows}{%
  \renewcommand{\arraystretch}{1.0}%
  \setlength{\aboverulesep}{1pt}%
  \setlength{\belowrulesep}{1.5pt}}
\newcommand{\pending}[1]{\textcolor{red}{[#1]}}
\newcommand{\sx}{\texttimes}
\newlength{\cellw}\newlength{\headw}\newlength{\labw}
\newcommand{\setcellw}[4]{%
  \settowidth{\cellw}{#1}%
  \settowidth{\headw}{#2}%
  \ifdim\headw>\cellw\setlength{\cellw}{\headw}\fi%
  \settowidth{\labw}{GLM-5.3-Flash\quad DFlash$^\dagger$}%
  \setlength{\headw}{\dimexpr(\textwidth-\labw-#4)/#3\relax}%
  \ifdim\headw>\cellw\setlength{\cellw}{\headw}\fi}
\newcommand{\hdr}[2]{\begin{tabular}[c]{@{}c@{}}#1\\#2\end{tabular}}
\providecommand{\resultrows}{\renewcommand{\arraystretch}{1.2}}
\newlength{\spdw}\newlength{\tauw}
\newcommand{\spdh}{\makebox[\spdw]{Spd.}}
\newcommand{\tauh}{\makebox[\tauw]{$\tau$}}
\newcommand{\nospd}{\makebox[\spdw]{--}}

\section{Experiments}
\label{sec:exp}

\subsection{Experimental setup}
\label{sec:exp:setup}

\paragraph{Models and baselines.}
We evaluate on three natively sparse targets, DS-V4.1-Flash~\citep{deepseekv4}, GLM-5.3-Flash~\citep{glm5}
and Hy4~flash~\citep{hy4flash}.
Their sparse attention differs in each case (\Cref{sec:bg}), so each sets its own paired layers and
top-$k$, the near window where it keeps one, and the corresponding number of positions per draft step
(\Cref{tab:targets}).
Every drafter has three layers of 128 routed experts, 6B to 14B parameters of which 0.4B to 1.1B are
active per token, and reuses the target's embedding and output head.
The baselines cover both ways of owning a cache (\Cref{sec:obs:own}): DFlash~\citep{dflash} keeps one
over the whole context, and DSpark~\citep{dspark} bounds its own to a sliding window.
DFlash$^\dagger$ is DFlash held to that same window, 128 tokens and 512 on Hy4~flash, which separates
the window from DSpark's other changes.
We train all three ourselves with the same recipe, data and number of steps as DFly2.
They have the same three-layer MoE stack and come within 2\% of its size.
Unless noted otherwise, every number we report is measured on NVIDIA H20 GPUs.

\begin{table*}[t]
\centering
\small
\caption{\textbf{What each target settles.} Positions per step is the top-$k$, the near window and the
block's six rows. A dash marks a window the top-$k$ already covers.}
\label{tab:targets}
\vspace{4pt}
\resultrows
\begin{tabular}{@{}lcrrr@{}}
\toprule
\textbf{Target} & \textbf{Paired layers $L$} & \textbf{Top-$k$} & \textbf{Near window} & \textbf{Positions per step} \\
\midrule
DS-V4.1-Flash & $[28, 32, 36]$    & 512                 & 128         & 646   \\
GLM-5.3-Flash       & $[35, 39, 43]$    & 512 pools of four   & --          & 2{,}054 \\
Hy4~flash           & $[33, 37, 41]$    & 2{,}048             & --          & 2{,}054 \\
\bottomrule
\end{tabular}
\end{table*}

\paragraph{Benchmarks.}
The three workload families, all taken from earlier work, differ in how much context a request carries.
The general-purpose family is the twelve evaluation sets that ship with DeepSpec, the training repository
released with DSpark~\citep{dspark}.
The main tables report seven of them: GSM8K~\citep{gsm8k}, MATH-500~\citep{math500},
AIME25~\citep{aime}, MBPP~\citep{mbpp}, HumanEval~\citep{humaneval},
LiveCodeBench~\citep{livecodebench} (LCB) and MT-Bench~\citep{mtbench}.
The long-context family follows LongSpark~\citep{longspark}: continuation tasks from LongSpec's
corpus~\citep{longspec} at 32K, LongSpark's own CodeSpan at 64K, and
LongSWE-Bench~\citep{longcodebench} at 128K, LSWE 128K in the tables.
The agentic family follows DLite~\citep{dlite}: AgentLongBench~\citep{agentlongbench} (ALB),
MemoryAgentBench~\citep{memoryagentbench} (MemAgent), the long-context multi-turn category of
BFCL-v3~\citep{bfcl}, and LongSWE-Bench issues at 64K (SWE 64K).
Each family runs under four sampling settings, greedy ($T = 0$) and stochastic ($T = 1$, or $T = 0.9$ on
Hy4~flash) crossed with 2K chat and 8K thinking.
The main tables report the 2K chat setting, thinking disabled and at most 2048 generated tokens.
The ablation tables average each family over its sets and all four settings, twelve family-setting
cells per drafter.

\paragraph{Training data.}
We train on DFly-SFT, a general-purpose instruction-tuning corpus of 989,861 conversations and 7.42B
tokens, all of it from public datasets.
The conversations fall into ten capability domains (\Cref{tab:data-mixture}), labeled by their source
where it gives a label and by an LLM annotator otherwise.
Samples are capped at 32,768 tokens, and each domain is filled with its longest samples first, which
makes the mixture long-form and reasoning-heavy.
An explicit reasoning trace appears in 61.6\% of samples and 34.9\% run past 8K tokens, so a drafter
meets long contexts during training.
The corpus is deduplicated globally and decontaminated against the twelve DeepSpec evaluation
sets~\citep{dspark} with an n-gram overlap test.

\begin{table*}[t]
\centering
\small
\caption{\textbf{DFly-SFT data mix.} Mix is the share of samples. Token counts and lengths use the
GLM-5.3-Flash tokenizer, and Think is the fraction of samples whose assistant turns carry an explicit
reasoning trace.}
\label{tab:data-mixture}
\vspace{4pt}
\resultrows
\begin{tabular}{@{}lrrrrr@{}}
\toprule
\textbf{Data type}    & \textbf{Mix} & \textbf{Samples} & \textbf{Tokens} & \textbf{Median len.} & \textbf{Think} \\
\midrule
Math (advanced)       & 19.2\%  & 190,000 & 2.90B  & 13,982      & 98.6\% \\
Code (competitive)    & 11.1\%  & 110,000 & 1.96B  & 18,175      & 94.7\% \\
Code (engineering)    & 9.6\%   & 95,479  & 0.56B  & 4,064       & 46.9\% \\
Knowledge \& STEM     & 15.2\%  & 150,000 & 0.48B  & 2,075       & 51.3\% \\
Instruction following & 7.1\%   & 70,000  & 0.47B  & 5,751       & 85.7\% \\
Code (function-level) & 11.1\%  & 110,000 & 0.41B  & 686         & 37.9\% \\
Writing               & 10.1\%  & 100,000 & 0.24B  & 1,425       & 40.1\% \\
Text transformation   & 6.1\%   & 60,000  & 0.19B  & 1,972       & 37.4\% \\
Logical reasoning     & 2.5\%   & 24,382  & 0.11B  & 1,461       & 65.0\% \\
Math (word problems)  & 8.1\%   & 80,000  & 0.09B  & 477         & 20.2\% \\
\midrule
Total                 & 100.0\% & 989,861 & 7.42B  & 3,621       & 61.6\% \\
\bottomrule
\end{tabular}
\end{table*}

\paragraph{Implementation details.}
The same recipe trains every drafter, on all three targets and in every ablation: three epochs at a
global batch of 256, sequence length 32K, with Muon~\citep{muon}, on the objective of \Cref{eq:loss}.
All speculative methods share a block of $\gamma = 5$ and run on vLLM~\citep{vllm}, where DFly2
registers no KV group and reads the pages its paired layers already hold.
Accepted length is measured on a single tp8 engine.
Each accepted length is the mean of the last three runs.
Speedups come from each target's own deployment at a concurrency of one, prefill and decode on separate
tp8 instances for DS-V4.1-Flash and GLM-5.3-Flash and one tp8 instance for Hy4~flash.
The concurrency sweep of \Cref{fig:serving} runs DS-V4.1-Flash greedily, every instance at DP8 $\times$
TP1 with EP8.
Short text pools the twelve general-purpose sets on one prefill and one decode instance, at one to 128
concurrent requests.
CodeSpan 64K and SWE 64K run at one to 32 with eight prefill instances per decode instance, since at
that length a single prefill instance cannot keep the decode batch fed.

\subsection{Experimental results}
\label{sec:exp:results}

\paragraph{General-purpose workloads.}
DFly2 accepts more than both baselines in all eighteen cells, and on the general-purpose family it is
3.9\% to 10.5\% above DSpark, the stronger baseline there (\Cref{tab:main-general}).
On short text, how much a drafter keeps in its own cache makes little difference, as holding DFlash to a
128-token window, 512 on Hy4~flash, moves its accepted length by at most 2\%.
DFly2 learns in training to draft from the positions the target's sparse attention selects, and trained
on its near window alone instead, it accepts less on this family for all three targets
(\Cref{tab:sharing}).
Training on the target's sparse selection thus helps the drafter generalize.

\begin{table*}[t]
\centering
\small
\resultrows
\setlength{\tabcolsep}{3pt}
\caption{Mean accepted length $\tau$ on general-purpose benchmarks, each cell at temperature 0/1
(0.9 on Hy4~flash). Bold and underline mark the best and second best per target and temperature.}
\label{tab:main-general}
\vspace{4pt}
\setcellw{\textbf{0.00}/\textbf{0.00}}{HumanEval}{8}{60pt}
\resizebox{\textwidth}{!}{%
\begin{tabular}{@{}ll@{\hspace{9pt}}*{3}{w{c}{\cellw}}@{\hspace{9pt}}*{3}{w{c}{\cellw}}@{\hspace{9pt}}w{c}{\cellw}@{\hspace{9pt}}w{c}{\cellw}@{}}
\toprule
 &  & \multicolumn{3}{c}{\textbf{Math}} & \multicolumn{3}{c}{\textbf{Code}} & \multicolumn{1}{c}{\textbf{Chat}} & \multicolumn{1}{c}{\textbf{Overall}} \\
\cmidrule(lr){3-5}\cmidrule(lr){6-8}\cmidrule(lr){9-9}\cmidrule(lr){10-10}
Target & Drafter & GSM8K & MATH & AIME25 & MBPP & HumanEval & LCB & MT-Bench & Avg. \\
\midrule
DS-V4.1-Flash & DFlash           &                  3.95/3.64 &                  3.93/3.42 &                  3.54/2.97 &                  3.82/3.63 &                  4.25/3.99 &                  3.34/2.80 &                  2.70/2.51 &                  3.65/3.28 \\
           & DFlash$^\dagger$ &                  3.94/3.59 &                  3.91/3.40 &                  3.51/2.94 &                  3.81/3.58 &                  4.08/3.82 &                  3.26/2.77 &                  2.68/2.53 &                  3.60/3.23 \\
           & DSpark           & \underline{4.24}/\underline{4.00} & \underline{4.18}/\underline{3.74} & \underline{3.69}/\underline{3.24} & \underline{4.00}/\underline{3.83} & \underline{4.27}/\underline{4.07} & \underline{3.44}/\underline{3.04} & \underline{2.82}/\underline{2.67} & \underline{3.81}/\underline{3.51} \\
\rowcolor{dflyrow}[3pt][3pt]
           & \textbf{DFly2}   & \textbf{4.62}/\textbf{4.45} & \textbf{4.43}/\textbf{4.10} & \textbf{4.01}/\textbf{3.60} & \textbf{4.28}/\textbf{4.17} & \textbf{4.70}/\textbf{4.56} & \textbf{3.76}/\textbf{3.38} & \textbf{3.03}/\textbf{2.91} & \textbf{4.12}/\textbf{3.88} \\
\cmidrule(l{0pt}r{0pt}){1-10}
GLM-5.3-Flash & DFlash           &                  3.93/3.65 &                  3.73/3.29 &                  3.25/2.77 &                  3.76/3.38 &                  3.84/3.44 &                  2.91/2.64 &                  2.58/2.34 &                  3.43/3.07 \\
           & DFlash$^\dagger$ &                  4.00/3.73 &                  3.83/3.38 &                  3.36/2.92 &                  3.79/3.43 &                  3.74/3.40 &                  2.99/2.73 &                  2.66/2.35 &                  3.48/3.13 \\
           & DSpark           & \underline{4.23}/\underline{3.97} & \underline{4.06}/\underline{3.61} & \underline{3.58}/\underline{3.17} & \underline{4.02}/\underline{3.66} & \underline{3.91}/\underline{3.62} & \underline{3.16}/\underline{2.92} & \underline{2.80}/\underline{2.50} & \underline{3.68}/\underline{3.35} \\
\rowcolor{dflyrow}[3pt][3pt]
           & \textbf{DFly2}   & \textbf{4.53}/\textbf{4.37} & \textbf{4.31}/\textbf{4.00} & \textbf{3.88}/\textbf{3.51} & \textbf{4.24}/\textbf{4.01} & \textbf{4.36}/\textbf{4.03} & \textbf{3.41}/\textbf{3.24} & \textbf{2.96}/\textbf{2.72} & \textbf{3.96}/\textbf{3.69} \\
\cmidrule(l{0pt}r{0pt}){1-10}
Hy4~flash  & DFlash           &                  3.86/3.77 &                  4.11/3.95 &                  3.80/3.69 &                  3.81/3.68 &                  3.02/3.04 &                  3.59/3.51 &                  2.90/2.71 &                  3.58/3.48 \\
           & DFlash$^\dagger$ &                  3.87/3.78 &                  4.13/3.98 &                  3.80/3.67 &                  3.83/3.69 &                  3.00/3.01 &                  3.61/3.52 &                  2.91/2.74 &                  3.59/3.48 \\
           & DSpark           & \underline{4.16}/\underline{4.09} & \underline{4.42}/\underline{4.29} & \underline{4.05}/\underline{3.98} & \underline{4.02}/\underline{3.92} & \underline{3.17}/\underline{3.22} & \underline{3.79}/\underline{3.72} & \underline{3.07}/\underline{2.92} & \underline{3.81}/\underline{3.73} \\
\rowcolor{dflyrow}[3pt][3pt]
           & \textbf{DFly2}   & \textbf{4.40}/\textbf{4.43} & \textbf{4.55}/\textbf{4.52} & \textbf{4.22}/\textbf{4.20} & \textbf{4.16}/\textbf{4.11} & \textbf{3.34}/\textbf{3.42} & \textbf{3.92}/\textbf{3.91} & \textbf{3.16}/\textbf{3.03} & \textbf{3.96}/\textbf{3.95} \\
\bottomrule
\end{tabular}%
}
\end{table*}

\paragraph{Long-context and agentic workloads.}
DFly2 holds its accepted length as prompts outgrow the 32K every drafter was trained at
(\Cref{tab:main-long}).
On the agentic family, whose prompts mostly run longer, its average stays within 4\% of its
general-purpose average on every target and temperature, while DSpark's falls 11\% to 21\%.
Its margin over the stronger baseline widens to 10.0\% to 31.5\% on the long-context and agentic families.
The baselines show the two limits of a drafter keeping its own context.
A window cannot see past itself, and a full cache sees everything but risks breaking where training
stops: on Hy4~flash, DFlash falls to as few as 1.08 tokens a step once prompts pass 32K.
DFly2 instead reads the target's KV at the positions the target's sparse indexer has already selected.
That selection can reach anywhere in the context, bounded by neither a near window nor the training
length, yet it stays a fixed top-$k$.
A draft step on a long prompt thus reads as many positions as it did in training, so what DFly2 learned
there carries over, and it inherits the target's length generalization.

\begin{table*}[t]
\centering
\small
\resultrows
\setlength{\tabcolsep}{3pt}
\caption{Mean accepted length $\tau$ on long-context and agentic benchmarks, formatted as in
\Cref{tab:main-general}. Long and Agen.\ average the three long-context and four agentic sets.}
\label{tab:main-long}
\vspace{4pt}
\setcellw{\textbf{0.00}/\textbf{0.00}}{MemAgent}{9}{63pt}
\resizebox{\textwidth}{!}{%
\begin{tabular}{@{}ll@{\hspace{9pt}}*{3}{w{c}{\cellw}}@{\hspace{9pt}}*{4}{w{c}{\cellw}}@{\hspace{9pt}}*{2}{w{c}{\cellw}}@{}}
\toprule
 &  & \multicolumn{3}{c}{\textbf{Long context}} & \multicolumn{4}{c}{\textbf{Agentic}} & \multicolumn{2}{c}{\textbf{Group average}} \\
\cmidrule(lr){3-5}\cmidrule(lr){6-9}\cmidrule(lr){10-11}
Target & Drafter & \hdr{LongSpec}{32K} & \hdr{CodeSpan}{64K} & \hdr{LSWE}{128K} & ALB & MemAgent & BFCL-v3 & \hdr{SWE}{64K} & Long & Agen. \\
\midrule
DS-V4.1-Flash & DFlash           & \underline{2.98}/\underline{2.55} & \underline{3.81}/\underline{3.13} &      \underline{3.42}/2.96 & \underline{4.38}/\underline{4.13} &                  2.84/2.60 & \underline{3.40}/\underline{3.11} & \underline{3.58}/\underline{3.14} & \underline{3.40}/\underline{2.88} & \underline{3.55}/\underline{3.25} \\
           & DFlash$^\dagger$ &                  2.74/2.40 &                  3.43/2.86 &                  3.15/2.77 &                  3.78/3.62 &                  2.80/2.52 &                  3.21/2.98 &                  3.19/2.80 &                  3.11/2.68 &                  3.24/2.98 \\
           & DSpark           &                  2.82/2.48 &                  3.58/2.92 &      3.29/\underline{2.98} &                  3.87/3.59 & \underline{2.85}/\underline{2.65} &                  3.33/3.09 &                  3.35/2.99 &                  3.23/2.79 &                  3.35/3.08 \\
\rowcolor{dflyrow}[3pt][3pt]
           & \textbf{DFly2}   & \textbf{3.12}/\textbf{2.81} & \textbf{4.14}/\textbf{3.59} & \textbf{3.97}/\textbf{3.55} & \textbf{5.10}/\textbf{4.72} & \textbf{3.36}/\textbf{3.13} & \textbf{3.77}/\textbf{3.55} & \textbf{4.02}/\textbf{3.60} & \textbf{3.74}/\textbf{3.32} & \textbf{4.06}/\textbf{3.75} \\
\cmidrule(l{0pt}r{0pt}){1-11}
GLM-5.3-Flash & DFlash           &                  2.55/2.17 &                  3.12/2.48 &                  2.92/2.64 &                  3.12/1.92 &      \underline{2.79}/2.50 &                  2.99/2.75 &                  2.91/2.60 &                  2.86/2.43 &                  2.95/2.44 \\
           & DFlash$^\dagger$ &                  2.69/2.23 &                  3.22/2.60 &                  2.97/2.65 &                  3.50/3.45 &                  2.68/2.48 &                  3.00/2.73 &                  2.86/2.61 &                  2.96/2.50 &                  3.01/2.82 \\
           & DSpark           & \underline{2.75}/\underline{2.37} & \underline{3.32}/\underline{2.77} & \underline{3.08}/\underline{2.79} & \underline{3.53}/\underline{3.73} &      2.74/\underline{2.58} & \underline{3.15}/\underline{2.91} & \underline{2.98}/\underline{2.74} & \underline{3.05}/\underline{2.64} & \underline{3.10}/\underline{2.99} \\
\rowcolor{dflyrow}[3pt][3pt]
           & \textbf{DFly2}   & \textbf{3.31}/\textbf{2.92} & \textbf{4.40}/\textbf{3.63} & \textbf{3.82}/\textbf{3.52} & \textbf{5.49}/\textbf{4.77} & \textbf{3.30}/\textbf{3.12} & \textbf{3.61}/\textbf{3.35} & \textbf{3.73}/\textbf{3.48} & \textbf{3.85}/\textbf{3.36} & \textbf{4.03}/\textbf{3.68} \\
\cmidrule(l{0pt}r{0pt}){1-11}
Hy4~flash  & DFlash           &                  2.59/2.41 &                  1.17/1.10 &                  1.12/1.09 &                  1.50/1.36 &                  1.08/1.06 & \underline{3.18}/\underline{3.20} &                  2.26/1.88 &                  1.63/1.53 &                  2.00/1.88 \\
           & DFlash$^\dagger$ &                  2.68/2.50 &                  3.47/3.28 &                  3.23/3.06 &      \underline{3.32}/3.06 &                  2.63/2.47 &                  2.90/2.92 &                  3.40/2.94 &                  3.13/2.95 &                  3.06/2.85 \\
           & DSpark           & \underline{2.74}/\underline{2.56} & \underline{3.54}/\underline{3.35} & \underline{3.31}/\underline{3.11} & \underline{3.32}/\underline{3.10} & \underline{2.77}/\underline{2.56} &                  3.08/2.98 & \underline{3.45}/\underline{3.14} & \underline{3.20}/\underline{3.01} & \underline{3.16}/\underline{2.95} \\
\rowcolor{dflyrow}[3pt][3pt]
           & \textbf{DFly2}   & \textbf{2.94}/\textbf{2.75} & \textbf{4.12}/\textbf{4.03} & \textbf{3.83}/\textbf{3.60} & \textbf{5.41}/\textbf{5.22} & \textbf{3.23}/\textbf{2.95} & \textbf{3.69}/\textbf{3.67} & \textbf{4.12}/\textbf{3.70} & \textbf{3.63}/\textbf{3.46} & \textbf{4.11}/\textbf{3.88} \\
\bottomrule
\end{tabular}%
}
\end{table*}

\subsection{Serving efficiency}
\label{sec:exp:serving}


\begin{table*}[t]
\centering
\small
\resultrows
\setlength{\tabcolsep}{3pt}
\caption{Speedup over the target alone on general-purpose benchmarks, greedy, at a concurrency of one.
Each cell gives decoding / end-to-end speedup in $\times$ over the target alone, the latter including
prefill. Bold and underline mark the
best and second best per target and speedup.}
\label{tab:speed-general}
\vspace{4pt}
\setcellw{\textbf{0.00}/\textbf{0.00}}{HumanEval}{8}{60pt}
\resizebox{\textwidth}{!}{%
\begin{tabular}{@{}ll@{\hspace{9pt}}*{3}{w{c}{\cellw}}@{\hspace{9pt}}*{3}{w{c}{\cellw}}@{\hspace{9pt}}w{c}{\cellw}@{\hspace{9pt}}w{c}{\cellw}@{}}
\toprule
 &  & \multicolumn{3}{c}{\textbf{Math}} & \multicolumn{3}{c}{\textbf{Code}} & \multicolumn{1}{c}{\textbf{Chat}} & \multicolumn{1}{c}{\textbf{Overall}} \\
\cmidrule(lr){3-5}\cmidrule(lr){6-8}\cmidrule(lr){9-9}\cmidrule(lr){10-10}
Target & Drafter & GSM8K & MATH & AIME25 & MBPP & HumanEval & LCB & MT-Bench & Avg. \\
\midrule
DS-V4.1-Flash & DFlash           & 3.23/2.87 & \underline{3.26}/\underline{3.14} & \underline{2.86}/2.73 & \underline{3.11}/\underline{3.04} & \underline{3.41}/\underline{3.29} & \underline{2.67}/\underline{2.54} & \underline{2.21}/\textbf{2.16} & \underline{2.96}/\underline{2.82} \\
           & DFlash$^\dagger$ & 3.21/2.83 & 3.19/3.06 & 2.75/2.71 & 3.09/3.02 & 3.28/3.17 & 2.62/2.49 & 2.19/2.14 & 2.91/2.77 \\
           & DSpark           & \underline{3.30}/\underline{2.91} & 3.24/3.11 & 2.80/\underline{2.76} & 3.09/3.02 & 3.31/3.18 & 2.63/2.52 & 2.15/2.11 & 2.93/2.80 \\
\rowcolor{dflyrow}[3pt][3pt]
           & \textbf{DFly2}   & \textbf{3.56}/\textbf{3.09} & \textbf{3.42}/\textbf{3.27} & \textbf{3.03}/\textbf{2.98} & \textbf{3.28}/\textbf{3.20} & \textbf{3.57}/\textbf{3.36} & \textbf{2.87}/\textbf{2.74} & \textbf{2.27}/\underline{2.15} & \textbf{3.14}/\textbf{2.97} \\
\cmidrule(l{0pt}r{0pt}){1-10}
GLM-5.3-Flash & DFlash           & 2.69/2.31 & 2.54/2.42 & 2.20/2.17 & 2.57/2.44 & 2.59/2.50 & 1.98/1.89 & 1.79/1.77 & 2.34/2.21 \\
           & DFlash$^\dagger$ & 2.77/2.39 & 2.62/2.50 & 2.26/2.22 & \underline{2.63}/2.49 & 2.56/2.47 & 2.04/1.98 & 1.83/1.80 & 2.39/2.26 \\
           & DSpark           & \underline{2.85}/\textbf{2.42} & \underline{2.67}/\underline{2.55} & \underline{2.32}/\underline{2.28} & \textbf{2.71}/\textbf{2.57} & \underline{2.62}/\underline{2.54} & \underline{2.10}/\underline{2.04} & \underline{1.85}/\underline{1.82} & \underline{2.45}/\underline{2.32} \\
\rowcolor{dflyrow}[3pt][3pt]
           & \textbf{DFly2}   & \textbf{2.90}/\underline{2.40} & \textbf{2.71}/\textbf{2.56} & \textbf{2.43}/\textbf{2.39} & \textbf{2.71}/\underline{2.56} & \textbf{2.74}/\textbf{2.64} & \textbf{2.16}/\textbf{2.10} & \textbf{1.88}/\textbf{1.85} & \textbf{2.50}/\textbf{2.36} \\
\cmidrule(l{0pt}r{0pt}){1-10}
Hy4~flash  & DFlash           & 3.23/2.98 & 3.47/3.35 & 3.21/3.16 & 3.21/3.09 & 2.55/2.50 & 3.03/2.84 & 2.44/2.39 & 3.02/2.90 \\
           & DFlash$^\dagger$ & 3.25/3.00 & 3.49/3.38 & 3.22/3.18 & 3.24/3.13 & 2.53/2.48 & 3.04/2.87 & 2.44/2.38 & 3.03/2.92 \\
           & DSpark           & \underline{3.47}/\underline{3.16} & \underline{3.71}/\underline{3.57} & \underline{3.42}/\underline{3.37} & \underline{3.38}/\underline{3.26} & \underline{2.67}/\underline{2.60} & \underline{3.17}/\underline{2.97} & \underline{2.58}/\underline{2.51} & \underline{3.20}/\underline{3.06} \\
\rowcolor{dflyrow}[3pt][3pt]
           & \textbf{DFly2}   & \textbf{3.70}/\textbf{3.34} & \textbf{3.85}/\textbf{3.69} & \textbf{3.56}/\textbf{3.51} & \textbf{3.51}/\textbf{3.38} & \textbf{2.81}/\textbf{2.75} & \textbf{3.30}/\textbf{3.07} & \textbf{2.66}/\textbf{2.59} & \textbf{3.34}/\textbf{3.19} \\
\bottomrule
\end{tabular}%
}
\end{table*}

\begin{table*}[t]
\centering
\small
\resultrows
\setlength{\tabcolsep}{3pt}
\caption{Speedup on long-context and agentic benchmarks, measured and formatted as in
\Cref{tab:speed-general}. DFlash$^\dagger$ reads a window at decode but still builds a full cache at
prefill.}
\label{tab:speed-long}
\vspace{4pt}
\setcellw{\textbf{0.00}/\textbf{0.00}}{MemAgent}{9}{63pt}
\resizebox{\textwidth}{!}{%
\begin{tabular}{@{}ll@{\hspace{9pt}}*{3}{w{c}{\cellw}}@{\hspace{9pt}}*{4}{w{c}{\cellw}}@{\hspace{9pt}}*{2}{w{c}{\cellw}}@{}}
\toprule
 &  & \multicolumn{3}{c}{\textbf{Long context}} & \multicolumn{4}{c}{\textbf{Agentic}} & \multicolumn{2}{c}{\textbf{Group average}} \\
\cmidrule(lr){3-5}\cmidrule(lr){6-9}\cmidrule(lr){10-11}
Target & Drafter & \hdr{LongSpec}{32K} & \hdr{CodeSpan}{64K} & \hdr{LSWE}{128K} & ALB & MemAgent & BFCL-v3 & \hdr{SWE}{64K} & Long & Agen. \\
\midrule
DS-V4.1-Flash & DFlash           & \textbf{2.46}/1.75 & \underline{3.05}/1.67 & \underline{2.59}/1.16 & \underline{3.46}/0.69 & 2.22/0.63 & \underline{2.67}/0.78 & \underline{2.79}/1.49 & \underline{2.70}/1.52 & \underline{2.78}/0.90 \\
           & DFlash$^\dagger$ & 2.32/1.70 & 2.70/1.55 & 2.52/1.13 & 3.09/0.83 & 2.25/0.69 & 2.63/0.84 & 2.55/1.47 & 2.51/1.46 & 2.63/0.96 \\
           & DSpark           & 2.28/\underline{1.99} & 2.63/\underline{2.06} & 2.49/\underline{1.68} & 2.95/\underline{1.21} & \underline{2.37}/\textbf{1.27} & 2.58/\textbf{1.37} & 2.54/\underline{1.89} & 2.47/\underline{1.91} & 2.61/\underline{1.44} \\
\rowcolor{dflyrow}[3pt][3pt]
           & \textbf{DFly2}   & \underline{2.44}/\textbf{2.10} & \textbf{3.23}/\textbf{2.37} & \textbf{2.94}/\textbf{1.86} & \textbf{3.82}/\textbf{1.25} & \textbf{2.60}/\underline{1.24} & \textbf{2.97}/\underline{1.36} & \textbf{3.00}/\textbf{2.13} & \textbf{2.87}/\textbf{2.11} & \textbf{3.10}/\textbf{1.50} \\
\cmidrule(l{0pt}r{0pt}){1-11}
GLM-5.3-Flash & DFlash           & 1.70/1.47 & 2.03/1.46 & 1.89/1.28 & 2.15/\underline{1.09} & 1.83/\textbf{1.01} & 2.02/\underline{1.16} & 1.88/1.36 & 1.87/1.40 & 1.97/1.15 \\
           & DFlash$^\dagger$ & \underline{1.86}/\underline{1.56} & 2.12/1.48 & 2.04/1.28 & 2.45/1.07 & \underline{1.92}/0.97 & 2.08/1.09 & 1.99/\underline{1.50} & 2.00/1.44 & 2.11/1.16 \\
           & DSpark           & 1.84/1.53 & \underline{2.24}/\underline{1.58} & \underline{2.12}/\underline{1.38} & \underline{2.48}/1.08 & 1.90/0.97 & \underline{2.17}/1.13 & \underline{2.02}/1.49 & \underline{2.07}/\underline{1.49} & \underline{2.14}/\underline{1.17} \\
\rowcolor{dflyrow}[3pt][3pt]
           & \textbf{DFly2}   & \textbf{2.14}/\textbf{1.75} & \textbf{2.82}/\textbf{1.79} & \textbf{2.40}/\textbf{1.39} & \textbf{3.57}/\textbf{1.10} & \textbf{2.20}/\underline{0.98} & \textbf{2.34}/\textbf{1.17} & \textbf{2.35}/\textbf{1.62} & \textbf{2.45}/\textbf{1.64} & \textbf{2.61}/\textbf{1.22} \\
\cmidrule(l{0pt}r{0pt}){1-11}
Hy4~flash  & DFlash           & 2.15/1.93 & 0.96/0.96 & 0.90/1.00 & 1.23/1.06 & 0.87/0.97 & \underline{2.64}/\underline{1.87} & 1.87/1.46 & 1.34/1.30 & 1.65/1.34 \\
           & DFlash$^\dagger$ & 2.28/2.03 & 2.96/2.26 & 2.75/\underline{1.60} & \underline{2.81}/\underline{1.26} & 2.18/\underline{1.29} & 2.46/1.82 & 2.90/1.80 & 2.66/1.96 & 2.59/1.54 \\
           & DSpark           & \underline{2.32}/\underline{2.06} & \underline{3.00}/\underline{2.28} & \underline{2.79}/\underline{1.60} & 2.80/\underline{1.26} & \underline{2.23}/1.27 & 2.59/1.85 & \underline{2.92}/\underline{1.81} & \underline{2.70}/\underline{1.98} & \underline{2.63}/\underline{1.55} \\
\rowcolor{dflyrow}[3pt][3pt]
           & \textbf{DFly2}   & \textbf{2.47}/\textbf{2.17} & \textbf{3.48}/\textbf{2.51} & \textbf{3.22}/\textbf{1.68} & \textbf{4.55}/\textbf{1.35} & \textbf{2.54}/\textbf{1.31} & \textbf{3.08}/\textbf{2.04} & \textbf{3.46}/\textbf{1.96} & \textbf{3.06}/\textbf{2.12} & \textbf{3.41}/\textbf{1.67} \\
\bottomrule
\end{tabular}%
}
\end{table*}

\paragraph{Throughput.}
DFly2 has the highest average speedup on every target and family, in decoding and end to end alike
(\Cref{tab:speed-general,tab:speed-long}).
Its lead in accepted length carries into decoding speed (\Cref{eq:sd-latency}) and, like it, is widest on the long-context and
agentic families.
Over the stronger baseline, DFly2 decodes 2\% to 6\% faster on the general-purpose family,
6\% to 18\% faster on the long-context one and 12\% to 30\% faster on the agentic one.
End to end, both DFlash variants on DS-V4.1-Flash pay for building their own cache over the prompt with
up to twice the time to the first token (\Cref{sec:obs:own}), and end the agentic sets slower than not
speculating.
DFly2 builds no cache of its own at prefill and thus leaves prefill speed unaffected.
On long prompts its advantage therefore has two sources, faster decoding from its higher accepted length
and no added overhead at prefill.

\begin{figure*}[t]
\centering
\includegraphics[width=\linewidth]{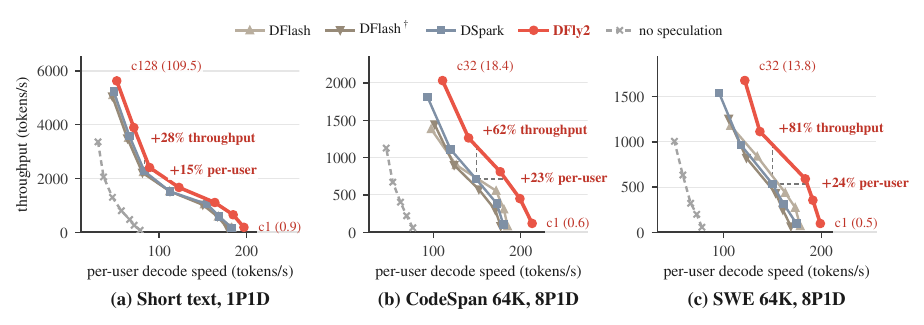}
\caption{Throughput against per-user decode speed on DS-V4.1-Flash, greedy, one point per concurrency
level. Labels give the concurrency and, in parentheses, the requests decoding. Dashed guides compare DFly2
with DSpark at equal per-user speed and at equal throughput. Throughput counts the whole deployment.}
\label{fig:serving}
\end{figure*}

\paragraph{High-concurrency throughput.}
We compare throughput and per-user decode speed across concurrency levels on short text and on a
long-context and an agentic set, CodeSpan 64K and SWE 64K (\Cref{fig:serving}).
DFly2's curve lies outside every baseline's at every level, and the gap does not close as the batch
grows.
At the same per-user speed as DSpark, DFly2 delivers 28\% higher throughput on short text and 62\% and
81\% higher on the two 64K sets, the largest gains again on long prompts.
At the same throughput, each user decodes 15\% to 24\% faster.
DFly2's advantage at a single request thus largely carries over to high concurrency.

\subsection{Ablation study}
\label{sec:exp:ablation}

\paragraph{What sharing buys.}
Of the two things the drafter shares, accepted length comes from the selection, whose scoring pass over
every position the target has already run (\Cref{tab:sharing,sec:obs:which}).
Removing its top-$k$ leaves only the near window, the largest loss here and two to six times as costly
on the long-context family as on the general-purpose one.
A window chooses by position, and the longer the context, the more it misses.
The near window itself is worth about 3\% on DS-V4.1-Flash and nothing on the other two.
On DS-V4.1-Flash the indexer scores one set of latents shared across layers, while each layer also
caches its own for the most recent 128 positions, which the top-$k$ cannot reach (\Cref{sec:method}).
A drafter with its own cache over that same near window accepts at most 3\% more on GLM-5.3-Flash and
Hy4~flash.
Sharing the cache therefore affects accepted length only slightly while removing the three costs of
\Cref{sec:obs:own}.
Because the drafter reads the selection as each target makes it, how much each term adds follows the
target's structure, and one design serves all three with nothing to tune in what it reads
(\Cref{sec:method}).

\begin{table*}[t]
\centering
\small
\setlength{\tabcolsep}{4pt}
\caption{Mean accepted length $\tau$ with one part of what DFly2 reads removed, averaged over each family's
sets and four sampling settings. Bold marks the best per column.}
\label{tab:sharing}
\vspace{4pt}
\tightrows
\begin{tabular}{@{}l>{\hspace{10pt}}rrr>{\hspace{10pt}}rrr>{\hspace{10pt}}rrr@{}}
\toprule
& \multicolumn{3}{c}{\textbf{DS-V4.1-Flash}} & \multicolumn{3}{c}{\textbf{GLM-5.3-Flash}} & \multicolumn{3}{c}{\textbf{Hy4~flash}} \\
\cmidrule(lr){2-4}\cmidrule(lr){5-7}\cmidrule(lr){8-10}
What the drafter reads & Gen. & Long & Agen. & Gen. & Long & Agen. & Gen. & Long & Agen. \\
\midrule
\rowcolor{dflyrow}[3pt][3pt]
\textbf{Full (DFly2)}            & \textbf{3.57} & \textbf{3.62} & \textbf{3.75} & 3.56 & 3.67 & \textbf{3.91} & \textbf{3.36} & \textbf{3.46} & 3.76 \\
w/o near window $\mathcal{N}$    & 3.45 & 3.50 & 3.63 & \textbf{3.57} & \textbf{3.68} & 3.86 & \textbf{3.36} & 3.45 & \textbf{3.77} \\
w/o the target's top-$k$         & 3.14 & 2.76 & 2.95 & 3.42 & 3.10 & 3.34 & 3.31 & 3.12 & 3.13 \\
w/o sharing (own cache)          & 3.46 & 3.34 & 3.37 & 3.46 & 3.18 & 3.43 & 3.32 & 3.14 & 3.17 \\
\bottomrule
\end{tabular}
\end{table*}
\begin{figure*}[t]
\centering
\includegraphics[width=\linewidth]{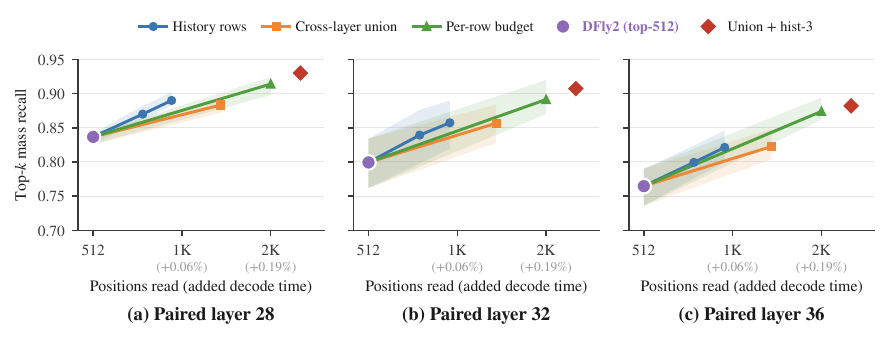}
\caption{Top-$k$ mass recall on DS-V4.1-Flash, the index-score share of each slot's own top-512 that a
read includes. Lines average slots 1--5 and three context lengths, with bands spanning the lengths.}
\label{fig:budget}
\end{figure*}

\begin{table*}[t]
\centering
\small
\begin{minipage}[t]{0.485\linewidth}
\caption{Mean accepted length $\tau$ on DS-V4.1-Flash, one change to DFly2 at a time, over each
family's sets and four sampling settings.}
\label{tab:ablation}
\end{minipage}\hfill
\begin{minipage}[t]{0.485\linewidth}
\caption{As in \Cref{tab:ablation}, with DFly2 reading beyond the target's top-$k$. Candidates reads
every first-stage candidate block, up to 16K positions.}
\label{tab:budget}
\end{minipage}

\vspace{4pt}
\setlength{\tabcolsep}{3pt}
\tightrows
\begin{minipage}[t]{0.485\linewidth}
\centering
\begin{tabular}[t]{@{}ll>{\hspace{2pt}}rrr@{}}
\toprule
\textbf{Dimension} & \textbf{Change} & \textbf{Gen.} & \textbf{Long} & \textbf{Agen.} \\
\midrule
--             & \textbf{DFly2}   & 3.57 & 3.62 & 3.75 \\
\grouprulepad
Pairing        & Adjacent 37--39  & 3.36 & 3.47 & 3.61 \\
\grouprulepad
Initialization & Attention only   & 3.51 & 3.56 & 3.69 \\
               & FFN only         & 3.51 & 3.58 & 3.72 \\
\grouprulepad
Correction     & w/o attn.\ reads & 3.47 & 3.51 & 3.65 \\
\grouprulepad
Injection      & w/o injection    & 3.51 & 3.56 & 3.68 \\
               & Broadcast        & \textbf{3.58} & \textbf{3.64} & \textbf{3.77} \\
\bottomrule
\end{tabular}
\end{minipage}\hfill
\begin{minipage}[t]{0.485\linewidth}
\centering
\begin{tabular}[t]{@{}ll>{\hspace{2pt}}rrr@{}}
\toprule
\textbf{Axis widened} & \textbf{Change} & \textbf{Gen.} & \textbf{Long} & \textbf{Agen.} \\
\midrule
--              & \textbf{DFly2}   & 3.57 & 3.62 & 3.75 \\
\grouprule
History rows    & hist-2           & 3.53 & 3.63 & 3.76 \\
                & hist-3           & 3.57 & 3.67 & 3.79 \\
\grouprule
Per-row budget  & top-1024         & 3.57 & 3.68 & 3.79 \\
\grouprule
Cross-layer     & Union            & 3.57 & 3.68 & 3.81 \\
\grouprule
Both            & Union+hist-3     & 3.57 & 3.71 & \textbf{3.84} \\
\grouprule
Selection stage & Candidates       & \textbf{3.58} & \textbf{3.72} & 3.82 \\
\bottomrule
\end{tabular}
\end{minipage}
\end{table*}

\paragraph{Which layers to pair with.}
The layers the drafter pairs with decide how many views of the context its three layers read.
Moving the pairing from layers 28, 32 and 36 to the adjacent layers 37 to 39 costs 4.6\% and lowers all
twelve cells (\Cref{tab:ablation}).
On DS-V4.1-Flash layers 28, 32 and 36 each run an indexer, while layers 37 to 39 reuse the 512 latents
layer 36 picked, so the draft layers read one view three times.
On a target that shares one cache across layers, the layers to pair with are thus those with an indexer
of their own.
Each of them contributes its own top-$k$, so the drafter gathers a richer picture of where the target
attends.

\paragraph{Where the weights start.}
Both halves of a draft layer start from its paired layer, the attention copied from it and the
feed-forward side drawn from its experts (\Cref{eq:attn-init,eq:ffn-init}), and each half helps about
equally.
Copying the attention alone, with the feed-forward side random, costs 1.6\%, and the reverse costs 1.1\%
(\Cref{tab:ablation}).
The copied weights come from the same layer whose latents the draft layer reads.
One pairing thus gives a draft layer, at no cost, both the context it reads and weights already matched
to it, aligning the draft layer with the target from the first step of training.

\paragraph{What the drafter adds.}
Both parts the drafter adds on top of what it shares raise accepted length.
Removing the two attention reads of the intra-block correction (\Cref{eq:correction}) costs 2.8\%, and removing the last-layer
hidden injection costs 1.6\% (\Cref{tab:ablation}).
The two reads take under 1\% of a decode step on a tp8 engine.
The correction lets each slot look back at the context and at the slots before it, and the injection
supplies the last-layer state from which the target produced the anchor.
Feeding the injection to every slot instead performs on par, so one entry point suffices.
Sharing hands the drafter the target's cache, and these two parts add what no cache holds, the block in
progress and the state behind the anchor.

\subsection{Analysis}
\label{sec:exp:analysis}

\paragraph{What YOCO opens.}
DS-V4.1-Flash follows YOCO~\citep{yoco} in keeping one set of latents that all layers share, and the
drafter reads that same cache.
Every indexer of the target therefore selects over what the drafter reads, which opens a design space in
how the drafter picks its top-$k$.
We explore three dimensions, earlier rows, more layers and a larger budget per row, and score each read
by its top-$k$ mass recall (\Cref{fig:budget}).
All three start from DFly2's own read, the target's top-512 at the row before the anchor, which on
average already recalls 0.80 of each slot's own top-512 by index score.
Adding two earlier rows reaches 0.86 at under 1K positions, the union of the target's five indexers
0.85 at 1.4K, and a 2K budget per row 0.89.
Earlier rows thus gain the most recall per added position, about twice the other two, and combining them
with the union reaches 0.91 at 2.6K positions.
Each read stays cheap, as 2K positions add 0.19\% to a decode step, against 14\% to 23\% for the whole
context at 128K (\Cref{sec:obs:own}).

\label{sec:analysis:budget}
Capped at 2K positions, the YOCO union with history matches the accepted length of reading every
first-stage candidate block, up to eight times as many (\Cref{tab:budget}).
On the general-purpose sets, contexts during drafting average under 2K positions, so the reads differ little in
what they see and accepted length rises by at most 0.3\%, comparable to the variation between runs.
On the long-context and agentic families, the union raises accepted length over DFly2's own read by 2.6\% and
2.4\%, against 2.7\% and 1.9\% for the candidate read.
Up to 16K positions, the target's top-$k$ thus holds nearly all that the drafter can use, and the
dimensions YOCO opens recover the rest at little cost.


\providecommand{\resultrows}{\renewcommand{\arraystretch}{1.2}}

\section{Related Work}
\label{sec:related}


\paragraph{Block-parallel drafters.}
Block-parallel drafters build a context KV of their own that spans the prompt and grows with it.
DFlash~\citep{dflash} fuses hidden states from several target layers, injects them into every draft layer
as extra keys and values, and drafts a whole block in one forward pass.
Its successors refine the conditioning signal~\citep{dflare,hyperdflash} or model the dependency between
slots~\citep{dspark,treeflash,domino}, and DFly~\citep{angelspec} does both.
All of them keep this cache, and their intra-block corrections do not return to the context.
DLite~\citep{dlite} avoids the cost by keeping no cache at all and conditioning on hidden states from many
target layers at the latest position.
DFly2 also builds no context KV of its own, but it still reads the context through the target's cache,
and so does its intra-block correction.

\paragraph{Reading the target's KV cache.}
Reading the target's KV cache instead of building one goes back to GliDe with CAPE~\citep{glide}, but
drafters on this line choose positions by rules that ignore the current query.
GliDe, LongSpec~\citep{longspec}, KVShot~\citep{kvshot} and Orthrus~\citep{orthrus} read the whole cache,
H-Spec~\citep{hspec} reads a fixed window, and LongSpark~\citep{longspark} reads a recent window plus a
summary pooled by fixed queries.
All of them were developed on dense targets, which attend to every position and so offer no selection to
follow.
DFly2 also reads the target's cache, but at the positions a natively sparse target attends to.

\paragraph{Query-aware selection.}
Query-aware selection gives good drafts, but prior work either finds it too costly for drafting or extracts
it from a dense target's attention.
TriForce~\citep{triforce} drafts from a cache retrieved by the current query, whereas
MagicDec~\citep{magicdec} finds per-query search too costly at large batch sizes and drafts from a static
sparse cache.
Verification-guided methods read the selection off the target's attention during verification.
Either the target itself then drafts~\citep{sparsespec,vegas}, or a separate drafter keeps a cache of its own
and reads only the chunks the target ranks highest~\citep{specextend}.
In every case, the drafting side has to compute or extract the selection itself.

\paragraph{Speculative decoding on natively sparse targets.}
Natively sparse targets such as DeepSeek-V3.2, DeepSeek-V4 and GLM-5 compute a query-aware selection as part
of their own attention~\citep{nsa,deepseekv32,deepseekv4,glm5}, and within a model one selection can already
serve several layers~\citep{indexcache}.
Systems work on these targets speeds up the target's side of speculative decoding, through sparse
verification kernels~\citep{ssv}, indexer prediction~\citep{spin} and tree verification over compressed
attention~\citep{treespecv4}.
The drafters paired with these targets, however, still bring a context of their own: the MTP modules shipped
with these models keep their own KV cache~\citep{glm5,dlite}, and DSpark serves DeepSeek-V4 with its own
128-token window~\citep{dspark}.
GLM-5.2 goes as far as reusing a selection across draft steps, but the selection is its MTP module's
own.\footnote{Enabled by \texttt{index\_share\_for\_mtp\_iteration} in the released GLM-5.2 configuration
(\url{https://huggingface.co/zai-org/GLM-5.2}). The sharing mechanism follows IndexCache~\citep{indexcache}.}
To our knowledge, no prior drafter reuses the target's own selection.
DFly2 reuses both the target's cache and its selection, so the drafter keeps no cache and computes no
selection of its own.

%

\section{Conclusion}
\label{sec:conclusion}

Long contexts have turned targets natively sparse, and speculative decoding increasingly runs on them.
Their drafters still bring a context of their own, a cache that costs more than the memory it takes,
read at positions a rule fixed before the query arrives.
A natively sparse target picks positions for each query already, over a cache it keeps anyway, and has
paid for both.
DFly2 shares both, reading each paired target layer's cached KV (\emph{KV sharing}) at the positions
that layer's indexer selected (\emph{Index sharing}).
It stores nothing per context position, scores nothing, and reads the same number of positions at any
prompt length.

On three targets whose sparse attention and architecture differ, DFly2 accepts more tokens per step
than DFlash and DSpark trained on the same recipe.
Its margin over the stronger one widens from 3.9\%--10.5\% on general workloads to 10.0\%--31.5\% on
long-context and agentic workloads of up to 128K.
At high concurrency on DS-V4.1-Flash, at matched per-user decode speed, that becomes 62\% and 81\% more
throughput at 64K.
The gain grows with the context because what DFly2 reads does not, the target's selection being a fixed
size that can still fall anywhere.
Long contexts are where serving needs speculative decoding most, and they are where following the
target's own choices pays most.

\newpage

\bibliographystyle{citation}
\bibliography{citation}

\newpage
\appendix

\end{document}